%% file: main.tex
\documentclass{article}

     \PassOptionsToPackage{numbers, compress}{natbib}

\usepackage[preprint]{neurips_2022}

\usepackage[utf8]{inputenc} 
\usepackage[T1]{fontenc}    
\usepackage{hyperref}       
\usepackage{booktabs}       
\usepackage{amsfonts}       
\usepackage{nicefrac}       
\usepackage{microtype}      

\usepackage{amsmath,amssymb,amsthm}
\usepackage{enumitem}
\usepackage[dvipnames]{xcolor}
\usepackage{url}
\usepackage{mathtools}
\usepackage{bbm}

\newtheorem{assumption}{Assumption}
\newtheorem{proposition}{Proposition}
\newtheorem{definition}{Definition}
\newtheorem{lemma}{Lemma}

\newtheorem{algorithms}{Algorithm}
\newtheorem{thm}{Theorem}

\newcommand{\parameter}{\omega}

\newcommand{\med}{\parameter}

\usepackage{wrapfig}
\usepackage{subfigure}
\usepackage{algorithm2e} 

\title{Adaptive Data Debiasing through\\
Bounded Exploration}

\author{%
        Yifan Yang \\ 
        Ohio State University \\
        \texttt{yang.5483@osu.edu} \\
        \And Yang Liu \\ 
        University of California, Santa Cruz \\
        \texttt{yangliu@ucsc.edu} \\
        \And Parinaz Naghizadeh \\
        Ohio State University \\
        \texttt{naghizadeh.1@osu.edu} \\  
}

\begin{document}

\maketitle
\vspace{-0.05in}
\begin{abstract}
Biases in existing datasets used to train algorithmic decision rules can raise ethical and economic concerns due to the resulting disparate treatment of different groups. 
We propose an algorithm for sequentially debiasing such datasets through adaptive and bounded exploration in a classification problem with costly and censored feedback. 
Exploration in this context means that at times, and to a judiciously-chosen extent, the decision maker deviates from its (current) loss-minimizing rule, and instead accepts some individuals that would otherwise be rejected, so as to reduce statistical data biases. Our proposed algorithm includes parameters that can be used to balance between the ultimate goal of removing data biases -- which will in turn lead to more accurate and fair decisions, and the exploration risks incurred to achieve this goal. We analytically show that such exploration can help debias data in certain distributions. We further investigate how fairness criteria can work in conjunction with our data debiasing algorithm. We illustrate the performance of our algorithm using experiments on synthetic and real-world datasets.
\end{abstract}

\input{Introduction}
\input{model}
\input{single-group}
\input{two-groups}
\input{simulations}

\newpage

\begin{small}
\bibliographystyle{abbrvnat}
\bibliography{fairness-bias}
\end{small}

\appendix
\section*{Appendix}
\input{supplement}

\end{document}

%% file: introduction.tex
\section{Introduction}\label{sec:intro}

Data-driven algorithmic decision making is being adopted widely to aid humans' decisions, in applications ranging from loan approvals to determining recidivism in courts. Despite their ability to process vast amounts of data and make accurate predictions, these algorithms can also exhibit and amplify existing social biases (e.g., \citep{dressel2018accuracy,lambrecht2019algorithmic,obermeyer2019dissecting}). There are at least two possible sources of unfairness in algorithmic decision rules: (data) biases in the training datasets, and (prediction) biases arising from the algorithm's decisions~\citep{mehrabi2019survey}. The latter problem has been receiving increasing attention, and is often addressed by imposing fairness constraints on the algorithm. In contrast, in this paper, we are primarily focused on the former problem of statistical biases in the training data itself. 

The datasets used for training machine learning algorithms might not accurately represent the agents they make decisions on, due to, e.g., historical biases in decision making and feature selection, or changes in the populations' characteristics or participation rates since the data was initially collected. 
Such data biases in turn can result in disparate treatment of underrepresented or disadvantaged groups; i.e., data bias can cause prediction/model bias, as also verified by recent work~\cite{kallus2018residual,wang2021fair,zhu2021rich,liao2022social}. Motivated by this, we focus on data biases, and propose an algorithm which, while attempting to make accurate (and fair) decisions, also aims to collect data in a way that helps it recover unbiased estimates of the characteristics of agents interacting with it. 

In particular, we study a classification problems with \emph{censored and costly feedback}. Censored feedback means that the decision maker only observes the true qualification state of those individuals it admits (e.g., a bank will only observe whether an individual defaults on or repays a loan if the loan is extended in the first place; an employer only assesses the performance of applicants it hires). In such settings, any mismatch between the available training data and the true population may grow over time due to adaptive sampling bias: once a decision rule is adopted based on the current training data, the algorithm's decisions will impact new data collected in the future, in that only agents passing the requirements set by the current decision rule will be admitted going forward. In response, the decision maker may attempt to collect more data from the population; however, such data collection is costly (e.g., in the previous examples, may require extending loans to/hiring unqualified individuals). Given these challenges, we present an \emph{active debiasing} algorithm with \emph{bounded exploration}: our algorithm admits some agents that would otherwise be rejected (i.e., it explores), yet adaptively and judiciously limits the extent and frequency of this exploration. 

Formally, consider a population of agents with features $x$, true qualification/labels $y$, and group memberships $g$ based on their demographic features. To design a (fair) algorithm that can minimize classification loss, the decision maker (implicitly) relies on estimates $\hat{f}^y_{g,t}(x)$ of the feature-label distribution of agents from group $g$, obtained from the current training dataset $\mathcal{H}_t=\{(x_{n},y_{n},g_{n})\}_{n=1}^{N_t}$. However, the resulting assumed distribution $\hat{f}^y_{g,t}(x)$ may be different from the true underlying distribution ${f}^y_g(x)$; this is the statistical data bias issue we focus on herein. Specifically, we consider distribution shifts between the estimates and the true distributions (Assumption~\ref{as:distributions}). 

\textbf{Our algorithm.} We propose an \texttt{active debiasing} algorithm (Algorithm~\ref{def:db-alg}), which actively adjusts its decisions with the goal of ensuring unbiased estimates of the underlying distributions ${f}^y_g(x)$ over time. In particular, at each time $t$, the algorithm selects a (fairness-constrained) decision rule that would minimize classification error based on its current, possibly biased estimates $\hat{f}^y_{g,t}(x)$; adopting this decision rule corresponds to \emph{exploitation} of the current information by the algorithm. At the same time, to circumvent the censored feedback nature of the problem, our algorithm  also deviates from the prescriptions of this loss-minimizing classifier to a judiciously chosen extent (the extent is chosen adaptively, based on the current estimates); this will constitute \emph{exploration}. Our algorithm includes two parameters to limit the costs of this exploration: one modulates the \emph{frequency} of exploration (an exploration probability $\epsilon_t$ {which can be adjusted using current bias estimates}), and another limits the \emph{depth} of exploration (by setting a threshold $\text{LB}_t$ on how far from the classifier one is willing to go when exploring). We show that these choices can strike a balance between the ultimate goal of removing statistical biases in the training data -- which will in turn lead to more accurate and fair decisions, and the cost of exploration incurred to achieve this goal. 

\textbf{Summary of findings and contributions.} 
Our main findings and contributions are as follows:

1. \emph{Comparison with baselines.}  We contrast our proposed algorithm against two baselines: an \texttt{exploitation-only} baseline (one that does not include any form of exploration), and a \texttt{pure exploration} baseline (which may randomly accept some of the agents rejected by the classifier, but does not bound exploration). We show (Theorem~\ref{thm:exploit-only}) that \texttt{exploitation-only} always leads to overestimates of the underlying distributions. Further, while \texttt{pure exploration} can debias the distribution estimates in the long-run (Theorem~\ref{thm:pure-exploration}), it does so at the expense of accepting \emph{any} agent, no matter how far from the classifier's threshold, leading to more costly exploration (Section~\ref{sec:simulations}).

2. \emph{Analytical support for our proposed algorithm.} We show (Theorem~\ref{thm:debiasing}) that our proposed \texttt{active debiasing} algorithm with bounded exploration can correct biases in {unimodal} distribution estimates. We also provide an error bound for our algorithm (Theorem~\ref{thm:regret}). 

3. \emph{Interplay with fairness criteria.} We analyze the impact of fairness constraints on our algorithm's performance, and show (Proposition~\ref{prop:fairness-debiasing}) that existing fairness criteria may speed up debiasing of the data in one group, while slowing it down for another. 

4. \emph{Numerical experiments.} We provide numerical support for the performance of our algorithm using experiments on synthetic and real-world (Adult and FICO) datasets.

\textbf{Related work.}
Our paper is closely related to the works of \citep{bechavod2019equal,kilbertus2020fair,ensign2018runaway,blum2020recovering,jiang2020identifying}, which study the impact of data biases on (fair) algorithmic decision making. Among these works, 
\citet{bechavod2019equal} and \citet{kilbertus2020fair} study \emph{fairness-constrained} learning in the presence of censored feedback. While these works also use exploration, the form and purpose of exploration is different: the algorithm in \citep{bechavod2019equal} starts with a pure exploration phase, and subsequently explores with the goal of ensuring the fairness constraint is not violated; the stochastic (or exploring) policies in \citep{kilbertus2020fair} conduct (pure) exploration to address the censored feedback issue. In contrast, we start with a {biased} dataset, and conduct \emph{bounded} exploration to debias data; fairness constraints may or may not be enforced separately and are orthogonal to our debiasing process. {Also, as shown in Section~\ref{sec:simulations}, such pure exploration processes can incur higher exploration costs than our proposed bounded exploration algorithm.} 

Our work is also closely related to \citep{deshpande2018accurate,nie2018adaptively,neel2018mitigating,wei2021decision}, which study adaptive sampling biases induced by a decision rule, particularly when feedback is censored. Among these, \citet{neel2018mitigating} also consider an \emph{adaptive} data gathering procedure, and show that no debiasing will be necessary if the data is collected through a differentially private method. We similarly propose an adaptive debiasing algorithm, but unlike \citep{neel2018mitigating}, account for the costs of exploration in our data collection procedure. The recent work of \citet{wei2021decision} studies data collection in the presence of censored feedback, and similar to our work, accounts for the cost of exploration in data collection, by formulating the problem as a partially observable Markov decision processes. Using dynamic programming methods, the data collection policy is shown to be a threshold policy that becomes more stringent (in our terminology, reduces exploration) as learning progresses. Our works are similar in that we both propose using adaptive and cost-sensitive exploration, but we differ in the problem setup and our analysis of the impact of fairness constraints. More importantly, in contrast to both \citep{neel2018mitigating,wei2021decision}, our starting point is a \emph{biased} dataset (which may be biased for reasons other than adaptive sampling in its collection, including historical biases); we then show how, while attempting to debias this dataset by collecting new data, any {additional} adaptive sampling bias during data collection can be prevented. 

Our work also falls within the fields of selective labeling bias, fair learning, and active learning. From the \emph{selective labeling bias} perspective, \citet{lakkaraju2017selective} propose a contraction technique to compare the performance of the predictive model and a human judge while they are forced to have the same acceptance rate. \citet{de2018learning} propose a data augmentation scheme by adding more samples that are more likely to be rejected (we refer to this as exploration) to correct the sample selection bias. From the \emph{fair learning} perspective, \citet{kallus2018residual} propose a re-weighting technique (re-weighing ideas are also explored in \citep{abernethy2020active, blum2020recovering, jiang2020identifying}) to solve the residual unfairness issue while accounting for adaptive sampling bias. From the \emph{active learning} perspective, \citet{noriega2019active} adaptively acquire additional information according to the needs of different groups or individuals given information budgets, to achieve fair classification. Similar to the approaches of these papers, we also compensate for adaptive sampling bias through exploration; the main difference, aside from the application, is in our analytical guarantees as well as our study of the interplay of data debiasing with fairness constraints. 

More broadly, our work has similarities to Bandit learning and its focus on exploration-exploitation trade-offs. A key difference of our work with existing bandit algorithms ($\epsilon$-greedy, UCB, EXP3, etc.) is our focus on \emph{bounded} exploration. We provide additional discussion on this, and review other related works \citep{perdomo2020performative,balcan2007margin,kazerouni2020active,liu2018delayed,liu2020disparate,zhang2019group,lakkaraju2017selective,de2018learning,kallus2018residual,noriega2019active,abernethy2020active} in more detail, in Appendix~\ref{app:related-additional}. 

%% file: model.tex
\vspace{-0.05in}
\section{Model and Preliminaries}\label{sec:model}
\vspace{-0.05in}
\textbf{The environment.} 
We consider a \emph{firm} or decision maker, who selects an {algorithm} to make decisions on a population of \emph{agents}. The firm observes agents arriving over times $t=1, 2, \ldots$ 
, makes a decision for agents arriving at time $t$ based on the current algorithm, and can subsequently adjust its algorithm for times $t+1$ onward based on the observed outcomes. 

Each agent has an observable \emph{feature} or \emph{score} $x\in \mathcal{X}\subseteq \mathbb{R}$.\footnote{We use a one-dimensional feature setting in our analysis, and generalize to $\mathcal{X}\subseteq \mathbb{R}^n$ in Section~\ref{sec:simulations}. Discussions and numerical experiments on potential loss of information due to our feature dimension reduction technique is given in Appendix~\ref{sec:extensions}.} These represent the agent characteristics that are leveraged by the firm in its decision; examples include credit scores or exam scores. Each agent is either qualified or unqualified to receive a favorable decision; this is captured by the agent's true \emph{label} or \emph{qualification state} $y\in \{0,1\}$, with $y=1$ and $y=0$ denoting qualified and unqualified agents, respectively. In addition, each agent in the population belongs to a different \emph{group} based on its demographic or protected attributes (e.g., race, gender); the agent's group membership is denoted $g\in \{a,b\}$. We consider threshold-based, group-specific, binary classifiers $h_{\theta_{g,t}}(x):\mathcal{X}\rightarrow \{0,1\}$ as (part of) the algorithm adopted by the firm, where $\theta_{g,t}$ denotes the classifier's decision threshold. An agent from group $g$ with feature $x$ arriving at time $t$ is admitted if $x\geq \theta_{g,t}$.

\textbf{Quantifying bias.} 
Let $f^y_{g}(x)$ denote the true underlying probability density function for the feature distribution of agents from group $g$ with qualification state $y$. The algorithm has an estimate of these unknown distributions, at each time $t$, based on the data collected so far (or an initial training set). Denote the algorithm's estimate at $t$ by $\hat{f}^y_{g,t}(x)$. In general, there can be a mismatch between the estimates $\hat{f}^y_{g,t}(x)$ and the true ${f}^y_{g}(x)$; this is what we refer to as bias. We assume the following. 

\begin{assumption}\label{as:distributions}
The firm updates its estimates $\hat{f}^y_{g,t}(x)$ by updating a single parameter $\hat{\parameter}^y_{g,t}$. 
\end{assumption}

This type of assumption is common in the multi-armed bandit learning literature \citep{schumann2019group,slivkins2019introduction,patil2021achieving,lattimore2020bandit,raab2021unintended} (there, the algorithm aims to learn the mean arm rewards). In our setting, it holds when the assumed underlying distribution is single-parameter, or when only one of the parameters of a multi-parameter distribution is unknown. Alternatively, it can be interpreted as identifying and correcting distribution shifts by updating a reference point in the distribution (e.g., adjusting the mean).\footnote{For instance, a bank may want to adjust for increases in average credit scores \citep{avg-CS-2,avg-CS} over time.} More specifically, we will let  $\hat{\parameter}^y_{g,t}$ be the $\alpha$-th percentile of $\hat{f}^y_{g,t}(x)$. {{We discuss potential limitations of Assumption~\ref{as:distributions} in Appendix~\ref{sec:extensions}}, and present an extension to a case with two unknown parameters in Appendix~\ref{app:two_parameters}.} 
 
Under Assumption~\ref{as:distributions}, the bias can be captured by the mismatch between the estimated and true parameters $\hat{\parameter}^y_{g,t}$ and ${\parameter}^y_{g}$. In particular, we set the mean absolute error $\mathbb{E}[|\hat{\parameter}^y_{g,t}-{\parameter}^y_{g}|]$ as the measure for quantifying bias, {where the randomness is due to that in $\hat{\parameter}^y_{g,t}$, the estimate of the unknown parameter based on data collected up to time $t$.}

\textbf{Algorithm choice without debiasing.} Let ${\alpha}^y_{g}$ be the fraction of group $g$ agents with label $y$. A loss-minimizing fair algorithm selects its thresholds $\theta_{g,t}$ at time $t$ as follows: 
\begin{align}
\min_{\theta_{a,t}, \theta_{b,t}} ~~ \sum_{g\in \{a,b\}} \alpha^1_{g}\int_{-\infty}^{\theta_{g,t}} \hat{f}^1_{g,t}(x)\mathrm{d}x + \alpha^0_{g}\int_{\theta_{g,t}}^{\infty} \hat{f}^0_{g,t}(x)\mathrm{d}x, \quad 
\text{s.t. } ~~ \mathcal{C}(\theta_{a,t}, \theta_{b,t}) = 0~.
    \label{eq:alg-obj}
\end{align}
Here, the objective is the misclassification error, and $\mathcal{C}(\theta_a, \theta_b) = 0$ is the fairness constraint imposed by the firm, if any. For instance, $\mathcal{C}(\theta_{a,t},  \theta_{b,t}) = \theta_{a,t} - \theta_{b,t}$ for \emph{same decision rule}, or  $\mathcal{C}(\theta_{a,t}, \theta_{b,t}) = \int_{\theta_{a,t}}^\infty 
\hat{f}^1_{a,t}(x)\mathrm{d}x - \int_{\theta_{b,t}}^\infty \hat{f}^1_{b,t}(x)\mathrm{d}x$ for \emph{equality of opportunity}. Note that both the objective function and the fairness constraint are affected by any inaccuracies in the current estimates $\hat{f}^y_{g,t}$. As such, a biased training dataset can lead to both loss of accuracy and loss in desired fairness. 

%% file: single-group.tex
\section{An \texttt{Active Debiasing} Algorithm with Bounded Exploration} 
\label{sec:single-group}

In this section, we present the \texttt{active debiasing} algorithm which uses both \emph{exploitation} (the decision rules of \eqref{eq:alg-obj}) and \emph{exploration} (some deviations) to remove any biases from the estimates $\hat{f}^y_{g,t}$. Although the deviations may lead to admission of some unqualified agents, they can be beneficial to the firm in the long-run: by reducing biases in $\hat{f}^y_{g,t}$, both classification loss estimates and fairness constraint evaluations can be improved. In this section, we drop the subscripts $g$ from the notation; when there are multiple groups, our algorithm can be applied to each group's estimates separately.

As noted in Section~\ref{sec:intro}, our algorithm is one of \emph{bounded} exploration: it includes a \emph{lower bound} $LB_t$, which captures the extent to which the decision maker is willing to deviate from the current classifier $\theta_t$, based on its current estimate $\hat{f}^0_t$ of the unqualified agents' underlying distribution. Formally, 
 
\begin{definition}\label{def:UB-LB}
At time $t$, the firm selects a lower bound $\text{LB}_t$ such that
\begin{align*}
    \text{LB}_t &= (\hat{F}^{0}_t)^{-1}(2\hat{F}^{0}_t(\hat{\med}^0_t)-\hat{F}^{0}_t(\theta_t)),
\end{align*}
where $\theta_t$ is the (current) loss-minimizing threshold determined from \eqref{eq:alg-obj}, $\hat{F}_t^0$, $(\hat{F}^{0}_t)^{-1}$ are the cdf and inverse cdf of the estimated distribution $\hat{f}_{t}^0$, respectively, and $\hat{\parameter}^{0}_t$ is (wlog) the $\alpha$-th percentile of $\hat{f}_{t}^0$. 
\end{definition}

In more detail, we choose $LB_t$ such that $\hat{F}^0_t(\hat{\omega}^0_t)-\hat{F}^0_t(LB_t)=\hat{F}^0_t(\theta_t)-\hat{F}^0_t(\hat{\omega}^0_t)$; that is, such that $\hat{\omega}^0_t$ is the median in the interval $[LB_t, \theta_t]$ based on the current estimate of the distribution $\hat{F}^0_t$ at the beginning of time $t$. Then, once a new batch of data is collected, we update $\hat{\omega}^0_t$ to $\hat{\omega}^0_{t+1}$, the \emph{realized} median of the distribution between $[LB_t, \theta_t]$ based on the data observed during $[t, t+1)$. Once the underlying distribution is correctly estimated, (in expectation) we will observe the same number of samples between $[LB_t, \hat{\omega}^0_t]$ and between $[\hat{\omega}^0_t, \theta_t]$, and hence $\hat{\omega}^0_t$ will no longer change. We also note that by selecting a high $\alpha$-th percentile in the above definition, $\text{LB}_t$ can be increased so as to limit the depth of exploration. As shown in Theorem~\ref{thm:debiasing}, and in our numerical experiments, these thresholding choice will enable debiasing of the distribution estimates while controlling its costs. 

Our \texttt{active debiasing} algorithm is summarized below. A pseudo-code is given in Appendix~\ref{app:pseudo}. 

\noindent\fbox{
    \parbox{0.93\textwidth}{
        \begin{algorithms}[The \texttt{active debiasing} algorithm]\label{def:db-alg}

Denote the loss-minimizing decision threshold determined from \eqref{eq:alg-obj} by $\theta_t$, and let $\text{LB}_t$ be given by Definition~\ref{def:UB-LB}. Let $\{\epsilon_t\}$ be a sequence of exploration probabilities. At each time $t$, and for agents $({x}^\dagger, {y}^\dagger)$ arriving at $t$: 

 \textbf{Step I: Admit agents and collect data.} Admit all agents with $x^{\dagger}\geq \theta_t$. Additionally,  if $\text{LB}_t\leq x^\dagger < \theta_t$, admit the agent with probability $\epsilon_t$.
 
 \textbf{Step II: Update the distribution estimates based on new data collected in Step I.}
\begin{itemize}[leftmargin=*]
\setlength\itemsep{0.01in}
    \item \textbf{Qualified agents' distribution update:} Identify new data with $\text{LB}_t \leq x^\dagger$ and ${y}^\dagger=1$. Use all such $x^\dagger$ with $\text{LB}_t \leq x^\dagger < \theta_t$, and such $x^\dagger$ with $\theta_t \leq x^\dagger$ with probability $\epsilon_t$, to update $\hat{\parameter}_t^1$. 
    \item \textbf{Unqualified agents' distribution update:} Identify new data with $\text{LB}_t \leq x^\dagger$ and ${y}^\dagger=0$. Use all such $x^\dagger$ with $\text{LB}_t \leq x^\dagger < \theta_t$, and such $x^\dagger$ with $\theta_t \leq x^\dagger$ with probability $\epsilon_t$, to update $\hat{\parameter}_t^0$. 
\end{itemize}
\end{algorithms}
    }
}

In more detail, our algorithm repeatedly performs the following two steps: 

\textbf{Step I: Data collection.} At the beginning of a time period $t$, a loss-minimizing classifier with threshold $\theta_t$ (according to \eqref{eq:alg-obj}) and the exploration lower bound $LB_t$ (Definition~\ref{def:UB-LB}) are selected based on the data collected so far. Then, given $\theta_t$, the new data collected during period $t$ will consist of arriving agents with features $x\geq \theta_{t}$. Additionally, to address the censored feedback issues, with probability $\epsilon_t$, the algorithm will also accept agents with ${{LB}_t} \leq x < \theta_t$. Note that this step balances between exploration and exploitation through its choice of both ${LB}_t$ (which limit the depth of exploration) and exploration probabilities $\epsilon_t$ (which limits the frequency of exploration).

\textbf{Step II: Updating estimates.} At the end of period $t$, the data collected in Step I will be used to update $\hat{f}^0_{t}$ and $\hat{f}^1_{t}$. Under Assumption~\ref{as:distributions}, the  estimates $\hat{f}^y_{t}$ are updated by updating the parameter $\hat{\parameter}^y_t$. We assume, without loss of generality, that the firm sets $\hat{\parameter}^{y}_t$ to the $\alpha$-th percentile of $\hat{f}_{t}^y$. This $\alpha$-th percentile is the \emph{reference point} that will be adjusted over time as new data is collected. As an example, when the reference point $\hat{\parameter}^{1}_t$ is set to the median (the 50-th percentile), the parameter can be adjusted so that half the label 1 data collected in Step I will lie on each side of the reference point. 

\section{Theoretical Analysis}\label{sec:analytical}

We begin by analyzing two baselines: \texttt{exploitation-only} (which only accepts agents with $x\geq \theta_t$, and uses no exploration or thresholding) and \texttt{pure exploration} (which accepts arriving agents at time $t$ who have $x<\theta_t$ with probability $\epsilon_t$, without setting any lower bound).
{The motivation for the choice of these two baselines is as follows: the \texttt{exploitation-only} baseline tracks the performance of a decision maker who is unaware of underlying data biases, and makes no attempt at fixing them. The \texttt{pure exploration} baseline, on the other hand, is motivated by the Bandit learning literature, and is also akin to debiasing algorithms proposed in recent work (see Section~\ref{sec:intro}, Related Work). We, in contrast, propose and show the benefits of bounded exploration through our \texttt{active debiasing} algorithm.}

\vspace{-0.05in}
\subsection{The {exploitation-only} baseline}
\vspace{-0.05in}
Our first baseline algorithm only updates its estimates of the underlying distributions based on agents with $x\geq \theta_t$ who pass the (current) loss-minimizing classifier \eqref{eq:alg-obj}. The following result shows that this approach consistently suffers from {adaptive sampling bias}, ultimately resulting in overestimation of the underlying distributions. 

\begin{thm}\label{thm:exploit-only}
An \texttt{exploitation-only} algorithm overestimates $\omega^y$, i.e., $\lim_{t\rightarrow \infty} \mathbb{E}[\hat{\parameter}^y_t]>\parameter^y, \forall y$. 
\end{thm}

A detailed proof is given in Appendix~\ref{app:proof_thm1}. 

\vspace{-0.05in}
\subsection{The {pure exploration} baseline}
\vspace{-0.05in}
In this second baseline, at each time $t$, the algorithm may accept any agent with $x<\theta_t$ with probability $\epsilon_t$. The following result establishes that using the data collected this way, the distributions can be debiased in the long-run, \emph{if} the data collected above the classifier is also sampled with probability $\epsilon_t$ when updating the distributions. 
\begin{thm}\label{thm:pure-exploration}
Using the \texttt{pure exploration} algorithm, $\hat{\parameter}^y_{t}\rightarrow {\parameter}^y$ as $t\rightarrow \infty$, $\forall y$. 
\end{thm}

The proof follows from assuming (wlog) that the unknown parameter $\parameter^y$ being estimated is the distribution's mean (can be generalized to arbitrary statistics under Assumption~\ref{as:distributions}). Then, as we are collecting i.i.d. samples from across the distribution, $\hat{\parameter}^y_t$ can be set to the sample mean of the collected data, and the conclusion follows from the strong law of large numbers. Note also that if \emph{all} the data above the classifier was considered when making the updates, following similar arguments to those in the proof of Theorem~\ref{thm:exploit-only}, the algorithm would obtain overestimates of the distributions. Lastly, we could equivalently balance data by resampling the exploration data (rather than downsampling the exploitation data), to debias data through this procedure. 

\vspace{-0.05in}
\subsection{The \texttt{active debiasing} algorithm}
\vspace{-0.05in}
While \texttt{pure exploration} can successfully debias data in the long-run, it does so at the expense of accepting agents with \emph{any} $x<\theta_t$. Below, we provide analytical support that our proposed exploration and thresholding procedure in the \texttt{active debiasing} algorithm can still debias data in certain distributions, while limiting the depth of exploration to $\text{LB}_t\leq x<\theta_t$. 

\begin{thm}\label{thm:debiasing}
Let  ${f}^y$ and $\hat{f}^y_{t}$ denote the true feature distribution and their estimates at the beginning of time $t$, with respective $\alpha$-th percentiles $\parameter^y$ and $\hat{\parameter}_t^y$. Assume these are unimodal distributions, $\epsilon_t>0, \forall t$, and {$\hat{\parameter}_t^0\leq\theta_t\leq \hat{\parameter}_t^1, \forall t$}.
Then, using the \texttt{active debiasing} algorithm,\\
\textbf{(a)} If $\hat{\parameter}^y_{t}$ is underestimated (resp. overestimated), then  $\mathbb{E}[\hat{\parameter}^y_{t+1}] \geq \hat{\parameter}_t^y,$ (resp. $\mathbb{E}[\hat{\parameter}^y_{t+1}] \leq \hat{\parameter}_t^y$) $\forall t, \forall y$.\\
\textbf{(b)} The sequence $\{\hat{\parameter}^y_{t}\}$ converges, with $\hat{\parameter}^y_{t}\rightarrow {\parameter}^y$ as $t\rightarrow \infty$, $\forall y$. 
\end{thm}

We provide a proof sketch for debiasing $\hat{f}_t^0$ which highlights the main technical challenges addressed in our analysis. The detailed proof is given in Appendix~\ref{app:debiasing_full}. 

\noindent\emph{Proof sketch:} Our proof involves the analysis of statistical estimates $\hat{\parameter}^0_{t}$ based on data collected from \emph{truncated} distributions. In particular, by bounding exploration, our algorithm will only collect data with features $x\geq \text{LB}_t$, and can use only this truncated data to build estimates of the unknown parameter of the distribution. 

Part \textbf{(a)} establishes that the sequence of $\{\hat{\parameter}^y_t\}$ produced by our \texttt{active debiasing} algorithm ``moves'' in the right direction over time, and ultimately converges. The main challenge in this analysis is that as the exploration and update intervals $[\text{LB}_t, \infty)$ are themselves adaptive, there is no guarantee on the number of samples in each interval, and therefore we need to analyze the estimates in finite sample regimes.
To proceed with the analysis, we assume the feature distribution estimates follow {unimodel} distributions (such as Gaussian, Beta, and the family of $alpha$-stable distributions) with $\parameter^0$ as reference points. We then consider the expected parameter update following the arrival of a batch of agents; Denote the current left portion in $(\text{LB}_t, \hat{\parameter}^0_t)$ as $p_1 := \frac{\hat{F}^0(\hat{\parameter}^0_t) -\hat{F}^0(\text{LB}_t)}{\hat{F}^0(\theta_t)-\hat{F}^0(\text{LB}_t)}$. Based on Definition ~\ref{def:UB-LB}, we can also obtain the current portion in $(\hat{\parameter}^0_t, \theta_t)$ denoted as $p_2 := \frac{\hat{F}^0(\theta_t) - \hat{F}^0(\hat{\parameter}^0_t)}{\hat{F}^0(\theta_t)-\hat{F}^0(\text{LB}_t)} = p_1$. The new expected estimates $\mathbb{E}[\hat{\parameter}^0_{t+1}]$ is the sample median in $(\text{LB}_t, \theta_t)$, where samples come from the true distribution. We establish that this expected update will be higher/lower than $\parameter^0_{t}$ if the current estimate is an under/over estimate of the true parameter. 

Then, in Part \textbf{(b)} we first show that the sequence of over- and under-estimation errors in $\{\hat{\parameter}^y_t\}$ relative to the true parameter ${\parameter}^y$ are supermartingales. By the Doobs Convergence theorem and using results from part \textbf{(a)}, these will converge to zero mean random variables with variance going to zero as the number of samples increases. This establishes that $\{\hat{\parameter}^y_t\}$ converges. It remains to show that this convergence point is the true parameter of the distribution. To do so, as detailed in the proof, we note that the density function of the sample median estimated on label 0 data collected in $[\text{LB}_t, \theta_t]$ is
\begin{align}
\mathbb P(\hat{\parameter}^0_{t}=\nu) \mathrm{d}\nu= \frac{(2m+1)!}{m!m!}(\tfrac{F^0(\nu)-F^0(\text{LB}_t)}{F^0(\theta_t)-F^0(\text{LB}_t)})^m(\tfrac{F^0(\theta_t)-F^0(\nu)}{F^0(\theta_t)-F^0(\text{LB}_t)})^m \tfrac{f^0(\nu)}{F^0(\theta_t)-F^0(\text{LB}_t)}\mathrm{d}\nu
\label{eq:median-pdf-trinomial}
\end{align}
which is a beta distribution {pushed forward} by $H(\nu):=\frac{F^0(\nu)-F^0(\text{LB}_t)}{F^0(\theta_t)-F^0(\text{LB}_t)}$; this is the CDF of the truncated $F^0$ distribution in $[\text{LB}_t, \theta_t]$. We then establish that the convergence point will be the true parameter of the underlying distribution. \hfill\qedsymbol

\vspace{-0.05in}
\subsection{Error bound analysis}\label{sec:regret-analysis}
\vspace{-0.05in}

Our error bound analysis compares the errors (measured as the number of wrong decisions made) of our \texttt{adaptive debiasing} algorithm against the errors that would be made by an oracle which knows the true underlying distributions. We measure the performance using 0-1 loss, $\ell(\hat{y_i},y_i)=\mathbbm{1}[\hat{y_i}\neq y_i]$, where $\hat{y}_i$ and $y_i$ denote the predicted and true label of agent $i$, respectively. We consider the error accumulated when updating the estimates using a total of $m$ batches of data. 
We split the total $T$ samples that have arrived during $[t, t+1)$ into four groups, corresponding to four different distributions ${f}^y_g$. Specifically, we use ${b}^y_{g,t}$ to denote the number of samples from each label-group pair at round $t \in \{0, \ldots, m\}$. We update the unknown distribution estimates once all batches meet a size requirement $s$, i.e, once $\min({b}^y_{g,t}) \geq s, \forall y, \forall g$. The error of our algorithm is given by:
\begin{align*}
    &\text{Error} = {\mathbb{E}}[Error_{Adaptive} - Error_{Oracle}] \\
    & = \sum_{t} \sum_{i=1}^{{b}^0_{a,t}+{b}^1_{a,t}+{b}^0_{b,t}+{b}^1_{b,t}} \mathop{\mathbb{E}}_{(x_i,y_i,g_i)\sim D}\Big[\ell(h_{\theta_{t,g}}(x_i,g_i),y_i)\Big] - \sum_{i=1}^{T} \displaystyle \mathop{\mathbb{E}}_{(x_i,y_i,g_i)\sim D}\Big[\ell(h^*_{\theta_g}(x_i,g_i),y_i)\Big] 
\end{align*}

The following theorem provides an upper bound on the error incurred by \texttt{active debiasing}. 
\begin{thm}~\label{thm:regret}
Let $\hat{f}^y_{g,t}(x)$ be the estimated feature-label distributions at round $t \in \{0, \ldots, m\}$. We consider the threshold-based, group-specific, binary classifier $h_{\theta_{g,t}}$, and denote the Rademacher complexity of the classifier family $\mathcal{H}$ with $n$ training samples by $\mathcal{R}_n(\mathcal{H})$. Let $\theta_{g,t}$ be a $v$-approximately optimal classifier based on data collected up to time $t$. At round $t$, let $N_{g,t}$ be the number of exploration errors incurred by our algorithm, $n_{g,t}$ be the sample size at time $t$ from group $g$, $d_{\mathcal{H}\Delta\mathcal{H}}(\tilde{D}_{g,t}, D_g)$ be the distance between the true unbiased data distribution $D_g$ and the current biased estimate $\tilde{D}_{g,t}$, and $c(\tilde{D}_{g,t}, D_g)$ be the minimum error on an algorithm trained on unbiased and biased data. Then, with probability at least $1 - 4\delta$ with $\delta >0$, the active debiasing algorithm's error is bounded by:
{\small \[
\text{Err.} \leq \sum_{g,t} \Big[ \underbrace{2v}_{\text{$v$-approx.}} + \underbrace{4\mathcal{R}_{n_{g,t}}(\mathcal{H}) + \tfrac{4}{\sqrt{n_{g,t}}} + \sqrt{\tfrac{2\ln(2/\delta)}{n_{g,t}}}}_{\text{empirical estimation errors}} + \underbrace{N_{g,t}}_{\text{explor.}} + \underbrace{d_{\mathcal{H}\Delta\mathcal{H}}(\tilde{D}_{g,t}, D_g)+2c(\tilde{D}_{g,t}, D_g)}_{\text{source-target distribution mismatch}} \Big]\]}
\end{thm}

More details on the definitions of the distance measure $d_{\mathcal{H}\Delta\mathcal{H}}$, and the error term $c(\cdot)$, and the exploration error term $N_{g,t}$, along with a a detailed proof, are given in Appendix~\ref{app:regret_proof}. From the expression above, we can see that the error incurred by our algorithm consists of four types of error: errors due to approximation of the optimal (fair) classifier at each round, empirical estimation errors, exploration errors, and errors due to our biased training data (viewed as source-target distribution mismatches); the latter two are specific to our \texttt{active debiasing} algorithm. In particular, as we collect more samples, $n_{g,t}$ will increase. Hence, the empirical estimation errors decrease over time. Moreover, as the mismatch between $\tilde{D}_{g,t}$ and $D_g$ decreases using our algorithm (by Theorem~\ref{thm:debiasing}), the error due to target domain and source domain mismatches also decrease. In the meantime, our exploration probability $\epsilon_t$ also becomes smaller over time, decreasing $N_{g,t}$. 

%% file: two-groups.tex
\vspace{-0.05in}
\subsection{Active debiasing and fairness criteria}\label{sec:two-group}
\vspace{-0.05in}
We next consider our proposed \texttt{active debiasing} algorithm when used in conjunction with demographic fairness constraints (e.g., equality of opportunity, same decision rule, and statistical parity \citep{mehrabi2019survey}). Imposing such fairness rules will lead to changes to the selected classifiers compared to the fairness-unconstrained case. Let $\theta^{F}_{g,t}$ and $\theta^{U}_{g,t}$ denote the fairness constrained and unconstrained decision rules obtained from \eqref{eq:alg-obj} at time $t$ for group $g$, respectively. We say group $g$ is being over-selected (resp. under-selected) following the introduction of fairness constraints if $\theta^{F}_{g,t}<\theta^{U}_{g,t}$ (resp. $\theta^{F}_{g,t}>\theta^{U}_{g,t}$). Below, we show how such over/under-selections can differently affect the debiasing of estimates on different agents. 

In particular, let the speed of debiasing be the rate at which  $\mathbb{E}[|\hat{\parameter}^y_t-\parameter^y|]$ decreases with respect to $t$; then, for a given $t$, an algorithm for which this error is larger has a slower speed of debiasing. The following proposition identifies the impacts of different fairness constraints on the speed of debiasing attained by our \texttt{active debiasing} algorithm. 
The proof is given in Appendix~\ref{app:fairness_debiasing}. 

\begin{proposition}\label{prop:fairness-debiasing}
Let  ${f}^y_g$ and $\hat{f}^y_{g,t}$ be the true and estimated feature distributions, with respective $\alpha$-th percentiles $\parameter^y$ and $\hat{\parameter}_t^y$. Assume these are unimodal distributions, and \texttt{active debiasing} is applied. 
If group $g$ is over-selected (resp. under-selected) under a fairness constraint, i.e.,  $\theta^{F}_{g,t}<\theta^{U}_{g,t}$ (resp. $\theta^{F}_{g,t}>\theta^{U}_{g,t}$), the speed of debiasing on the estimates $\hat{f}^y_{g,t}$ will decrease (resp. increase).  
\end{proposition}

Proposition~\ref{prop:fairness-debiasing} highlights the following implications of using both fairness rules and our active debiasing efforts. Some fairness constraints (such as equality of opportunity) can lead to an increase in opportunities for (here, over-selection of) agents from disadvantaged groups, while others (such as same decision rule) can lead to under-selection from that group. Proposition~\ref{prop:fairness-debiasing} shows that \texttt{active debiasing} may in turn become faster or slower at debiasing estimates on this group. 

Intuitively, over-selection provides increased opportunities to agents from a group (compared to an unconstrained classifier). In fact, the reduction of the decision threshold to $\theta^{F}_{g,t}$ can itself be interpreted as introducing exploration (which is separate from that introduced by our debiasing algorithm). When a group is over-selected under a fairness constraint, the fairness-constrained threshold $\theta^F_{g,t}$ will be lower than the unconstrained threshold $\theta^U_{g,t}$. Therefore, the exploration range will be narrower, which means by adding a fairness constraint, the algorithm needs to wait and collect more samples (takes a longer time) before it manages to collect sufficient data to accurately update the unknown distribution parameter, and hence, it has a slower debiasing speed. More broadly, these findings 
contribute to our understanding of how fairness constraints can have long-term implications beyond the commonly studied fairness-accuracy tradeoff when we consider their impacts on data collection and debiasing efforts. 

%% file: simulations.tex
\section{Numerical Experiments}\label{sec:simulations}

In this section, we illustrate the performance of our algorithm through numerical experiments on both Gaussian and Beta distributed synthetic datasets, and on two real-world datasets: the \emph{Adult} dataset~\citep{Dua:2019} and the \emph{FICO} credit score dataset~\citep{fico-data} pre-processed by~\citep{hardt2016equality}. Additional details (ground-truth information) on the experiments, and larger versions of all figures, are available in Appendix~\ref{app:figs}. Our code is available at: \url{https://github.com/Yifankevin/adaptive_data_debiasing}. 

Throughout, we either choose a \emph{fixed} schedule for reducing the exploration frequencies $\{\epsilon_t\}$, or reduce these \emph{adaptively} as a function of the estimated error. For the latter, the algorithm can select a range (e.g., above the classifier for label 0/1) and adjust the exploration frequency proportional to the discrepancy between the number of observed classification errors in this interval relative to the number expected given the distribution estimates. 

\begin{figure*}[ht]
\centering
	\subfigure[Rate of debiasing, $f^1$ and $f^0$ underestimated.]
	{
		\includegraphics[width=0.225\linewidth]{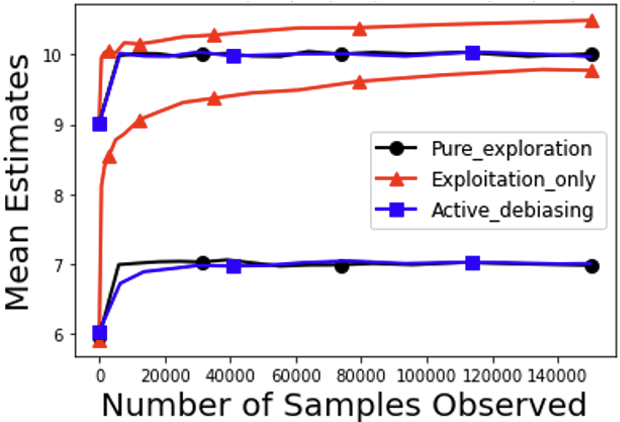}
		\label{fig:1under_0under_baseline}
	}
	\subfigure[Rate of debiasing, $f^1$ and $f^0$ overestimated.]
	{
		\includegraphics[width=0.225\textwidth]{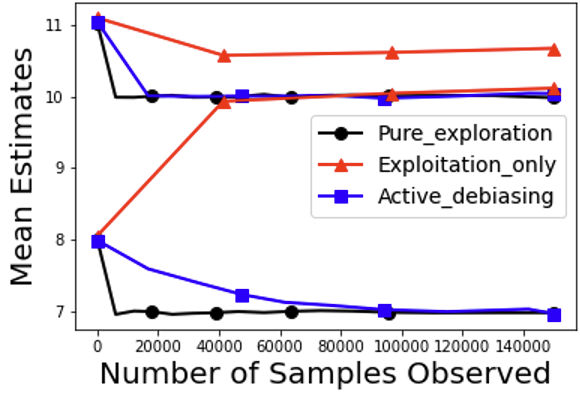}
		\label{fig:1under_0over_baseline}
	}
    \subfigure[Regret.]
    {
		\includegraphics[width=0.235\textwidth]{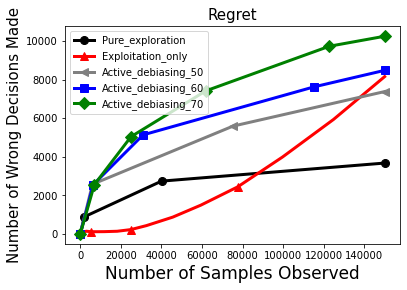}
		\label{fig:regret}
	}
    \subfigure[Weighted regret.]
    {
		\includegraphics[width=0.235\textwidth]{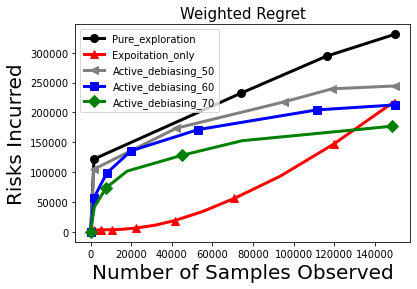}
		\label{fig:weighted-regret}
	}
\caption{Speed of debiasing, regret, and weighted regret, of \texttt{active debiasing} vs. \texttt{exploitation-only} and \texttt{pure exploration} (larger figures in Appendix~\ref{app:figs}).}
\label{fig:baseline}
\end{figure*}

\textbf{Comparison with the \texttt{exploitation-only} and \texttt{pure exploration} baselines:} Our first experiments in Fig.~\ref{fig:baseline}, compare our algorithm against two baselines. The underlying distributions are Gaussian and no fairness constraint is imposed. Our algorithm sets $\alpha^1=50$ and $\alpha^0=60$ percentiles, and exploration frequencies $\epsilon_t$ are selected adaptively by both our algorithm and {pure exploration}.

\underline{Speed of debiasing}: Figs.~\ref{fig:1under_0under_baseline} and~\ref{fig:1under_0over_baseline} show that consistent with Theorem~\ref{thm:exploit-only}, \texttt{exploitation-only} overestimates the distributions due to adaptive sampling biases. Further, consistent with Theorem~\ref{thm:pure-exploration}, \texttt{pure exploration} successfully debiases data. We also observe that as expected, \texttt{pure exploration} debiases \emph{faster} than \texttt{active debiasing}. The difference is more pronounced in the label 0 distributions compared to label 1, where \texttt{pure exploration} collects more ``diverse'' observations than our algorithm. For this same reason, the gap between \texttt{pure exploration} and our algorithm is larger when ${f}^0$ is overestimated. This is because \texttt{pure exploration} observes samples with \emph{lower} features $x$ than \texttt{active debiasing}, and so can use these to reduce its estimate faster. 

\underline{Regret}: Figs.~\ref{fig:regret} and~\ref{fig:weighted-regret} compare the regret and weighted regret of the algorithms. Regret is measured as the difference between the number of FN+FP decisions of an algorithm vs the oracle loss-minimizing algorithm derived on unbiased data. Formally, regret is defined as in Section~\ref{sec:regret-analysis}; weighted regret is defined similarly, but also adds a weight to each FN or FP decision, with the weight exponential in the distance of the feature of the admitted agent from the classifier. We observe that \texttt{exploitation-only}'s regret is super-linear, as not only it fails to debias, but has increasing error due to biases from overestimating. On the other hand, while algorithms that explore ``deeper'' have lower regret (e.g. \texttt{pure exploration} $<$ \texttt{active debiasing} with $\alpha^0=50 <$ \texttt{active debiasing} with $\alpha^0=60$ in Figs.~\ref{fig:regret}), they have higher weighted regret (the order is reversed in Fig.~\ref{fig:weighted-regret}). In other words, exploring to admit agents with low features $x$ leads to some errors, but ultimately helps reduce future mistakes, leading to sub-linear regret. However, if the risk/cost of these wrong decisions is taken into account, the firm may be better off adopting slower, but less risky exploration thresholds (e.g. $\alpha^0=70$). 

\begin{wrapfigure}[10]{r}{0.55\textwidth}\vspace{-0.2in}
	\centering
	\subfigure{
		\includegraphics[width=0.25\textwidth]{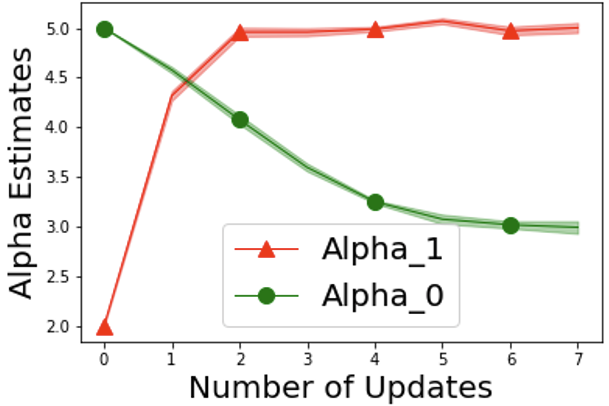}
	}
	\subfigure{
		\includegraphics[width=0.25\textwidth]{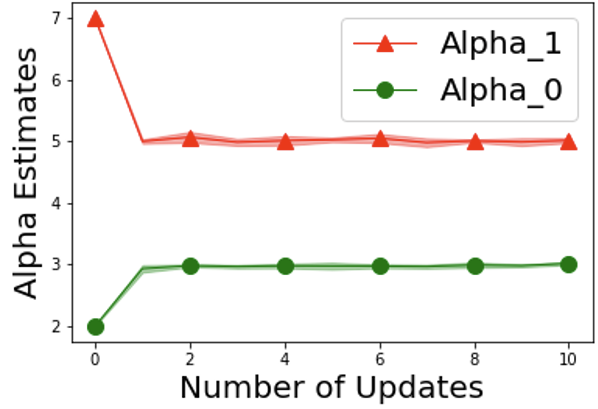}
	}
	\vspace{-0.1in}
	\caption{\texttt{Debiasing} under Beta distributions.}
	\label{fig:debiasing-beta}
\end{wrapfigure}
\textbf{Performance of \texttt{active debiasing} on Beta distributions:} Fig.~\ref{fig:debiasing-beta} shows that our algorithm can debias data for which the underlying feature-label distributions follow Beta distributions. We have assumed a mistmach between the parameter $\alpha$ of the true and estimated distributions, and selected these so that the estimated and true distributions have different relative skewness. This verifies that Theorem~\ref{thm:debiasing} holds for asymmetric distributions. 

\begin{wrapfigure}[7]{r}{0.55\textwidth}\vspace{-0.32in}
	\centering
	\vspace{-0.4in}
 	\subfigure[Advantaged label 0.]{
		\includegraphics[width=0.25\textwidth]{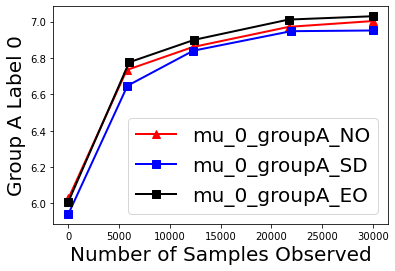}
		\label{fig:mu0_groupA}
	}
	\hspace{0.05in}
	\subfigure[Disadvantaged label 0.]{
		\includegraphics[width=0.25\textwidth]{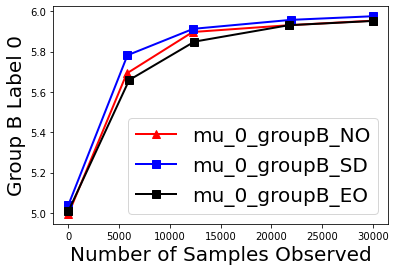}
		\label{fig:mu0_groupB}
	}
	\vspace{-0.1in}
	\caption{\texttt{Debiasing} used with fairness constraints.}
	\label{fig:fairness-debiasing}
\end{wrapfigure}
\textbf{Interplay of debiasing and fairness constraints:} Fig.~\ref{fig:fairness-debiasing} compares the performance of \texttt{active debiasing} when there are two groups of agents with underlying Gaussian distributions, and the algorithm is chosen subject to three different fairness settings: no fairness, equality of opportunity (EO), and the same decision rule (SD). The findings are consistent with Proposition~\ref{prop:fairness-debiasing}. {For instance, SD will over-select the majority group (i.e., $\theta^{SD}_{a,t} < \theta^U_{a,t}$) so that, as shown in the left panel in Fig.~\ref{fig:fairness-debiasing}, the speed of debiasing on the estimates $\hat{\omega}^0_{a,t}$ will decrease. In contrast, an opposite effect will happen in the minority group $b$ which is under-selected (i.e., $\theta^{SD}_{b,t} > \theta^U_{b,t}$). The effects of EO can be similarly explained by noting that it under-selects the majority group and over-selects the minority group.}

\begin{figure}[ht]
	\vspace{-0.1in}
	\centering
	\subfigure[Debiasing $G_a$, Adult.]
	{
		\includegraphics[width=0.23\textwidth]{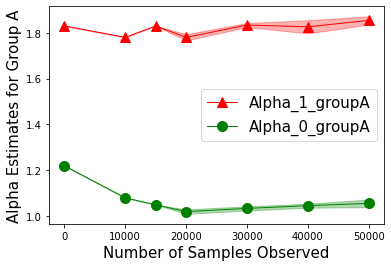}
		\label{fig:adult-white}
	}
	\subfigure[Debiasing $G_b$, Adult.]
	{
		\includegraphics[width=0.23\textwidth]{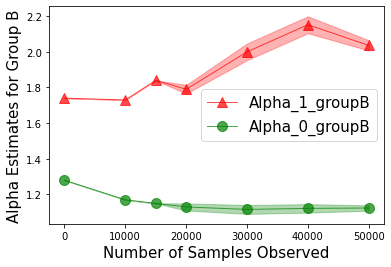}
		\label{fig:adult-nonwhite}
	}
	\subfigure[$G_b$, augmented with synthetic data (Adult).]
	{
		\includegraphics[width=0.23\textwidth]{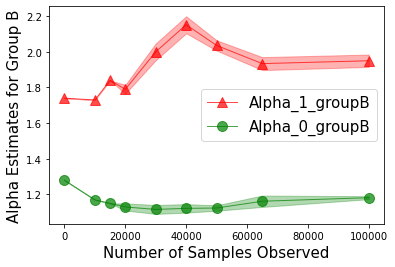}
		\label{fig:group-b-extra-data}
	}
	\hspace{0.02in}
	\subfigure[\texttt{Debiasing} on \emph{FICO}.]{
		    \includegraphics[width=0.23\textwidth]{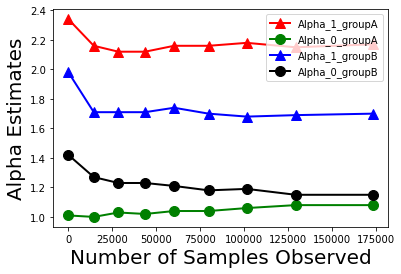}
		\label{fig:FICO_groupAandB}}
	\caption{\texttt{Active debiasing} on the \emph{Adult} and \emph{FICO} datasets.}
	\vspace{-0.2in}
	\label{fig:adult}
\end{figure}

\textbf{\texttt{Active debiasing} on the \emph{Adult} dataset:} Fig.~\ref{fig:adult} illustrates the performance of our algorithm on the \emph{Adult} dataset. Data is grouped based on race (White ${G}_a$ and non-White ${G}_b$), with labels $y=1$ for income $>\$50k/year$. A one-dimensional feature $x \in \mathbb{R}$ is constructed by conducting logistic regression on four quantitative and qualitative features (education number, sex, age, workclass), based on the initial training data.\footnote{While this experiment maintains the same mapping throughout, the mapping could be periodically revised.} Using an input analyzer, we found Beta distributions as the best fit to the underlying distributions. We use {2.5\%} of the data to obtain a biased estimate of the parameter $\alpha$. The remaining data arrives sequentially. We use $\alpha^1=50$ and $\alpha^0=60$ and a fixed decreasing $\{\epsilon_t\}$, with the equality of opportunity fairness constraint imposed throughout. 

We observe that our proposed algorithm can debias estimates across groups and for both labels,  but that this happens in the long-run and given access to sufficient samples. In particular, we note that for label 1 agents from $G_b$, as there are only {1080} samples in the dataset, although the bias initially decreases, the final estimate still differs from the true value. Fig.~\ref{fig:group-b-extra-data} verifies that this estimate would have been debiased in the long-run, had additional samples from the underlying population become available (i.e., as more such agents arrive). 

\textbf{\texttt{Active debiasing} on the \emph{FICO} dataset:} Fig.~\ref{fig:adult} also illustrates the performance of our algorithm on the \emph{FICO} dataset \cite{fico-data,hardt2016equality}, and shows that it is successful in debiasing distribution estimates on both groups and on both labels. 

\section{Conclusion, Limitations, and Future Work}

We proposed an \texttt{active debiasing} algorithm which recovers unbiased estimates of the underlying data distribution of agents interacting with it over time. We also analyzed the interplay of our proposed statistical/data debiasing effort with existing social/model debiasing efforts, shedding light on the potential alignments and conflicts between these two goals in fair algorithmic decision making. We further illustrated the performance of our proposed algorithm, and its interplay with fairness constraints, through numerical experiments on both synthetic and real-world datasets. 

\textbf{The single-unknown parameter assumption.} Our work focuses on learning of a single unknown parameter (Assumption~\ref{as:distributions}). Despite the commonality of this assumption in the multi-armed bandit learning literature, it also entails parametric knowledge of the underlying distribution with the other parameters such as variance or spread being known. We extend our algorithm to a Gaussian distribution with two unknown parameters in Appendix~\ref{app:two_parameters}. Extensions beyond this, especially those not requiring parametric assumptions on the underlying distributions, remain a main direction of future work. 

\textbf{On one-dimensional features and threshold classifiers.} Our analytical results have been focused on one-dimensional feature data and threshold classifiers. These assumptions may not be too restrictive in some cases: the optimality of threshold classifiers has been established in the literature by, e.g., \citep[Thm 3.2]{corbett2017algorithmic} and \citep{raab2021unintended}, as long as a multi-dimensional feature can be mapped to a properly defined scalar. Moreover, the recent advances in deep learning have helped enable this possibility: one can take the last layer outputs from a deep neural network and use it as the single dimensional representation. That said, any reduction of multi-dimensional features to a single-dimensional score may lead to some loss of information. In particular, our experiments have considered the use of our \texttt{active debiasing} algorithm on the \emph{Adult} dataset with multi-dimensional features by first performing a dimension reduction to a single-dimensional score; we find that this reduction can lead to a $\sim 5\%$ loss in performance (see Appendix~\ref{sec:extensions} for details). One potential solution to this is to adopt a mapping from high-dimensional features to scores that is revised repeatedly as the algorithm collects more data. Alternatively, one may envision a debiasing algorithm which targets its exploration towards collecting data on features that are believed to be highly biased; these remain as potential extensions of our algorithm. 

\textbf{Potential social impacts.} More broadly, while our debiasing algorithm imposes fairness constraints on its exploitation decisions (see problem~\eqref{eq:alg-obj}), it does not consider fairness constraints in its \emph{exploration} decisions. That means that our proposed algorithm could be disproportionate in the way it increases opportunities for qualified or unqualified agents in different groups during exploration. Imposing fairness rules on exploration decisions, as well as identifying algorithms that can improve the speed of debiasing of estimates on underrepresented populations, can be explored to address these potential social impacts, and remain as interesting directions of future work.  

Additional discussions on limitations, extensions, and social impacts, are given in Appendix~\ref{sec:extensions}.

\begin{ack}
We sincerely thank the three reviewers and the area chair for their comments and feedback which helped improve our paper. We are also grateful for support from the National Science Foundation (NSF) program on Fairness in AI in collaboration with Amazon under Award No. IIS-2040800, the NSF under grant IIS-2143895, and Cisco Research. Any opinion, findings, and conclusions or recommendations expressed in this material are those of the authors and do not necessarily reflect the views of the NSF, Amazon, or Cisco.
\end{ack}

%% file: supplement.tex
\input{discussion}

\input{related-additional}

\section{Pseudo-code for Algorithm~\ref{def:db-alg}}\label{app:pseudo}
The pseudo-code is shown in Algorithm~\ref{alg:two}. 
\input{pseudocode}

\section{Proof of Theorem~\ref{thm:exploit-only}}\label{app:proof_thm1}

\input{Proof_Thm1}

\section{Proof of Theorem~\ref{thm:debiasing}}\label{app:debiasing_full}

\input{full_proof_active_debiasing}

\section{Proof of Theorem~\ref{thm:regret}}\label{app:regret_proof}
\input{regret_proof}

\section{Proof of Proposition~\ref{prop:fairness-debiasing}}\label{app:fairness_debiasing}

\input{proof_fairness_debiasing}

\section{Larger figures and additional experiments}\label{app:figs}

\subsection{Larger figures and experiment details}

\begin{figure}[htbp!]
\centering
		\includegraphics[width=0.5\textwidth]{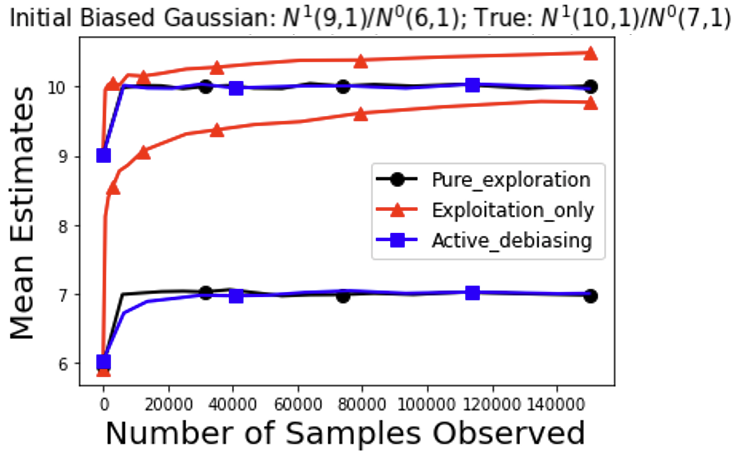}
		\label{fig:1under_0under_baseline_large}
		\caption{Rate of debiasing of our active debiasing algorithm vs the two baselines. Underlying feature distributions are Gaussian. We let $f^1$ be underestimated with $\hat{\parameter}^1=9$ and true parameter $\parameter^1=10$  (parameter debiasing shown in the top lines), and $f^0$ underestimated with $\hat{\parameter}^0=6$ and true parameter $\parameter^0=7$ (parameter debiasing shown in bottom lines).}
\end{figure}

\begin{figure}[htbp!]
\centering
		\includegraphics[width=0.5\textwidth]{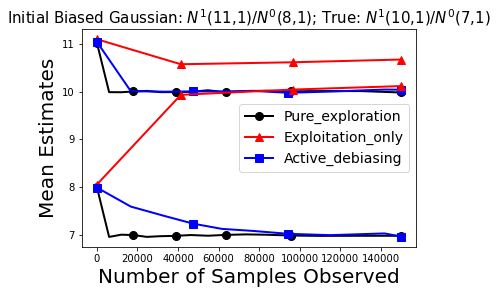}
		\label{fig:1under_0over_baseline_large}
		\caption{Rate of debiasing of our active debiasing algorithm vs the two baselines. Underlying feature distributions are Gaussian. We let $f^1$ be overestimated with $\hat{\parameter}^1=11$ and true parameter $\parameter^1=10$  (parameter debiasing shown in the top lines), and $f^0$ overestimated with $\hat{\parameter}^0=8$ and true parameter $\parameter^0=7$ (parameter debiasing shown in bottom lines).}
\end{figure}

\begin{figure}[htbp!]
\centering
		\includegraphics[width=0.4\textwidth]{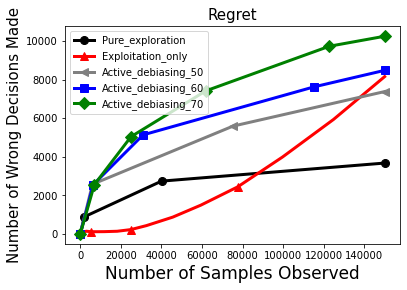}
		\label{fig:regret_large}
		\caption{Regret, measured as the number of wrong decisions (sum of False Positives and False Negatives) compared to an oracle classifier which knows the unbiased underlying distributions. Underlying feature distributions are Gaussian. We let $f^1$ and $f^0$ be overestimated with $\hat{\parameter}^1=9$ and $\hat{\parameter}^0=6$, and their true parameters $\parameter^1=10$ and $\parameter^0=7$, respectively.}
\end{figure}

\begin{figure}[htbp!]
\centering
		\includegraphics[width=0.4\textwidth]{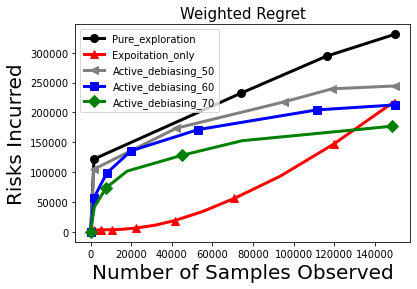}
		\label{fig:weighted-regret_large}
		\caption{Weighted regret, measured as the \emph{risk} of the wrong decisions made (a weighted sum of False Positives and False Negatives) compared to an oracle classifier which knows the unbiased underlying distributions. The weighted regret for each label 0 (resp. 1) sample is calculated by the exponential of the difference between its feature and the forth standard deviation above (resp. below) the mean. Underlying feature distributions are Gaussian. We let $f^1$ and $f^0$ be overestimated with $\hat{\parameter}^1=9$ and $\hat{\parameter}^0=6$, and their true parameters $\parameter^1=10$ and $\parameter^0=7$, respectively.}
\end{figure}

\begin{figure}[htbp!]
	\centering
 	\subfigure[Advantaged label 0. We assume We assume the assumed distribution is underestimated, with estimated parameter $\hat{\parameter}^0_a=6$ and true parameter $\parameter^0_a=7$.]
 	{
		\includegraphics[width=0.4\textwidth]{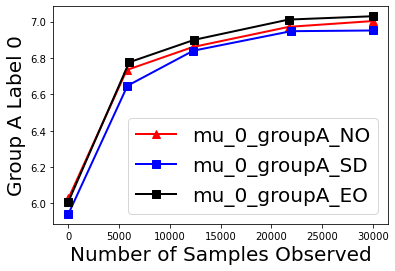}
		\label{fig:mu0_groupA_large}
	}
	\hspace{0.05in}%
	\subfigure[Disadvantaged label 1. We assume the assumed distribution is underestimated, with estimated parameter $\hat{\parameter}^0_b=8$ and true parameter $\parameter^0_b=9$.]
	{
		\includegraphics[width=0.4\textwidth]{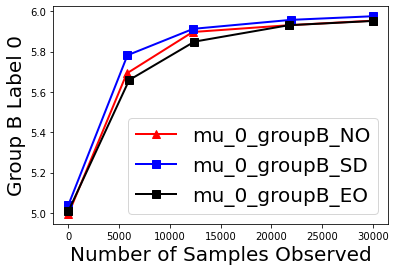}
		\label{fig:mu0_groupB_large}
	}
	\caption{\texttt{Active debiasing} under different fairness constraints. The underlying feature distributions are Gaussian, and the reference points are set to $\alpha^0=60$ and $\alpha^1=50$ for both groups. The exploration frequency $\{\epsilon_t\}$ is reduced with the fixed schedule of being subtracted by 0.1 after observing every 3000 samples.}
	\label{fig:fairness-debiasing_large}
\end{figure}

\begin{figure}[htbp!]
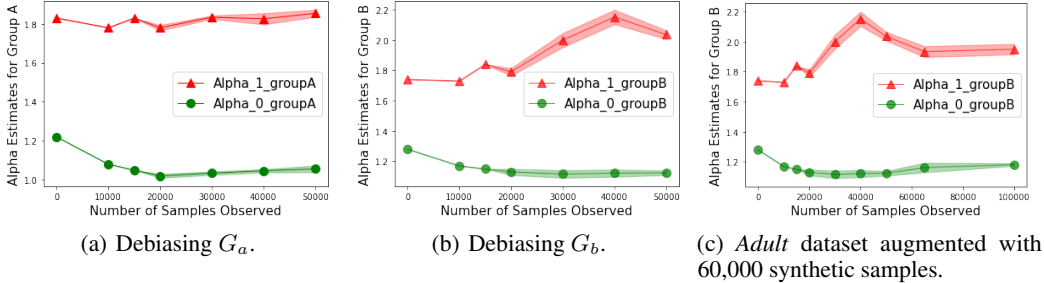

	\centering
	\subfigure[Debiasing $G_a$.]
	{
		\includegraphics[width=0.315\textwidth]{Figs/Adult_GroupA_FSR_and_bar.png}
		\label{fig:adult-white_large}
	}
	\subfigure[Debiasing $G_b$.]
	{
		\includegraphics[width=0.315\textwidth]{Figs/Adult_GroupB_FSR_and_bar.png}
		\label{fig:adult-nonwhite_large}
	}
	\subfigure[\emph{Adult} dataset augmented  with 60,000 synthetic samples.]
	{
		\includegraphics[width=0.315\textwidth]{Figs/Adult_GroupB_FSR_with_augmented_data_and_bar.png}
		\label{fig:group-b-extra-data_large}
	}
	\caption{Illustration of the performance of \texttt{active debiasing} on the \emph{Adult} dataset. The true underlying distributions were estimated to be Beta distributions with parameters Beta(1.94, 3.32) and Beta(1.13, 4.99) for group $a$ (White) label 1 and 0, respectively, and Beta(1.97, 3.53) and Beta(1.19, 6.10) for group $b$ (non-White) label 1 and 0, respectively. We used 2.5\% of the data to fit initial assumed distributions Beta(1.83, 3.32) and Beta(1.22, 4.99) for group $a$ label 1 and 0, respectively, and Beta(1.74, 3.53) and Beta(1.28, 6.10) for group $b$ label 1 and 0, respectively. The equal opportunity fairness constraint is imposed throughout. The exploration frequency $\{\epsilon_t\}$ is reduced with the fixed schedule of being subtracted by 0.1 after observing every 10000 samples}
	\label{fig:adult_large}
\end{figure}

\begin{figure}[htbp!]
	\centering
	\subfigure[Active Debiasing on the \emph{FICO} dataset.]
	{
		\includegraphics[width=0.4\textwidth]{Figs/FICO_Alpha_estimates_with_EO_and_FSR.png}
		\label{fig:FICO_groupAandB_large}
	}
	\hspace{0.1in}
	\subfigure[Difference w.r.t. the true value.]
	{
		\includegraphics[width=0.4\textwidth]{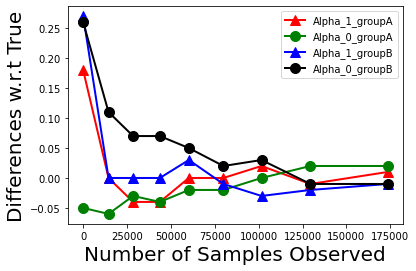}
		\label{fig:FICO_diff_large}
	}
	\caption{Illustration of the performance of \texttt{active debiasing} on the \emph{FICO} dataset. The true underlying distributions were estimated to be Beta distributions with parameters Beta(2.16, 1.27) and Beta(1.06, 3.98) for group $a$ (White) label 1 and 0, respectively, and Beta(1.71, 1.62) and Beta(1.16, 5.51) for group $b$ (non-White) label 1 and 0, respectively. We used 0.3\% of the data to fit initial assumed distributions Beta(2.34, 1.27) and Beta(1.01, 3.98) for group $a$ label 1 and 0, respectively, and Beta(1.98, 1.62) and Beta(1.42, 5.51) for group $b$ label 1 and 0, respectively. The equal opportunity fairness constraint is imposed throughout. The exploration frequency $\{\epsilon_t\}$ is reduced with the fixed schedule of being subtracted by 0.1 after observing every 17000 samples}
	\label{fig:fico_large}
\end{figure}

\subsection{Additional experiments: effects of depth of exploration}

\begin{figure}[htbp!]
	\centering
 	\subfigure[False positives (unqualified agents admitted) under each reference point]
 	{
		\includegraphics[width=0.475\textwidth]{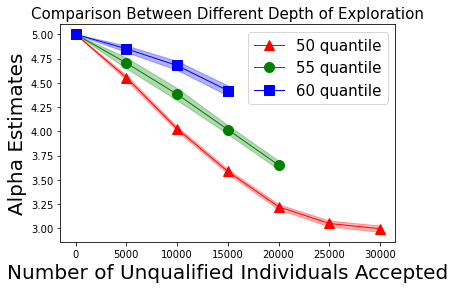}
		\label{fig:FP_reference-points}
	}
	\hspace{0.05in}
	\subfigure[False negatives (qualified agents rejected) under each reference point]
	{
		\includegraphics[width=0.475\textwidth]{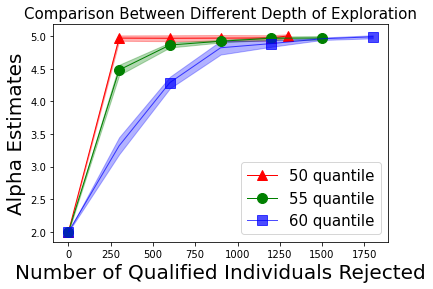}
		\label{fig:FN_reference-points}
	}
	\caption{\texttt{Active debiasing} under different choices of depth of exploration, with $\alpha^1=50$ and $\alpha^0=\{50,55,60\}$. We reduce $\{\epsilon_t\}$ following a fixed reduction schedule. The underlying feature distributions are Beta distributions.}
	\label{fig:reference-points}
\end{figure}

Figure~\ref{fig:reference-points} compares the effects of modifying the depth of exploration through the choice of reference points on the performance of our \texttt{active debiasing} algorithm. In particular, we fix $\alpha^1=50$ as the reference point on the qualified agents' estimates, and vary the reference points on unqualified agents' estimates in $\alpha^0\in\{50,55,60\}$, with smaller reference points indicating deeper exploration (see Definition~\ref{def:UB-LB}). In all three settings, we reduce $\{\epsilon_t\}$ following a fixed reduction schedule, as described in Section~\ref{sec:simulations}.

We first note that as also observed earlier, increasing the depth of exploration (here, e.g., setting $\alpha^0=50$) leads to faster speed of debiasing. This additional speed comes with a tradeoff: Fig.~\ref{fig:FP_reference-points} shows that algorithms with deeper exploration make more false positive errors, as they accept more unqualified individuals during exploration; by taking on this additional risk, they can debias the data faster. In addition, as observed in Fig.~\ref{fig:FN_reference-points}, the increased speed of debiasing means that the algorithm ultimately ends up making \emph{fewer} false negative decisions on the qualified individuals as a result of obtaining better estimates of their distributions. 

We conclude that a decision maker can use the choice of the reference point $\alpha^0$ in our proposed algorithm to achieve their preferred tradeoff between the risk incurred due to incorrect admissions (higher FP) vs the benefit from the increased speed of debiasing and fewer missed opportunities (fewer FN). 

\section{Debiasing with two unknown parameters: a Gaussian distribution with two unknown parameters mean $\mu$ and variance $\sigma^2$}\label{app:two_parameters}

In this section, we extend our algorithm to debias the estimates of distributions with two unknown parameters. Specifically, we consider a single group, and assume that the underlying feature-label distributions are Gaussian distributions for which both the mean and variance are potentially incorrectly estimated by the firm. 

We follow our \texttt{active debiasing} algorithm, with a choice of medians as reference points (i.e., $\alpha^i=50, \forall i$), and setting the thresholds UB (See Definition ~\ref{def:UB} below) and LB so that the reference points are the medians of the truncated distribution between the bounds and the classifier $\theta$. For this experiment only, we set a UB similar to LB for simplicity. We then follow Algorithm~\ref{def:db-alg}'s procedure with the same type of exploitation and exploration decisions, and with the additional step that now we update both parameters when updating the underlying estimates. 

\begin{definition}\label{def:UB}
At time $t$, the firm selects a upper bound $\text{UB}_t$ such that
\begin{align*}
    \text{UB}_t = (\hat{F}^{1}_t)^{-1}(2\hat{F}^{1}_t(\hat{\med}^1_t)-\hat{F}^{1}_t(\text{LB}_t)),
\end{align*}
where $\text{LB}_t$ is obtained from Definition \ref{def:UB-LB}, $\hat{F}_t^1$, $(\hat{F}^{1}_t)^{-1}$ are the cdf and inverse cdf of the estimated distribution $\hat{f}_{t}^1$, respectively, and $\hat{\parameter}^{1}_t$ is (wlog) the $\alpha$-th percentile of $\hat{f}_{t}^1$. 
\end{definition}

In order to update the mean and  variance estimates for obtaining $\hat{f}^i_{t}$, we find the sample mean and sample variance of the collected data, incrementally. However, we note that the obtained sample mean and sample variances are \emph{for truncated distributions}; the truncations are due to the presence of a classifier which limits the admission of a samples, as well as due to our proposed bounds UB and LB in the data collection procedure. We therefore need to convert between the estimated statistics for the truncated distribution and those of the full distribution accordingly. 

Specifically, we obtain the sample mean of the truncated distribution as follows: 
\begin{align*}
    \hat{\mu}^i_{t+1}=\frac{x_1+x_2+...+x_{n_i}+x^\dagger}{N^i_t+1} =  \frac{N^i_t}{N^i_t+1}\hat{\mu}^i_{t}+\frac{x^\dagger}{N^i_t+1}, \quad i \in \{0, 1\}~.
\end{align*}
where $N^i_t$ is the existing number of agents in the pool, and $\mu^i_t$ is the current (truncated) mean value estimate for label $i=\{0, 1\}$.

For the sample (truncated) variance for group $i$, $(\hat{s}^i_t)^2$, the updating procedure is:
\begin{align*}
(\hat{s}^i_{t+1})^2&=\frac{\sum_{j=1}^{N^i_t}(\hat{\mu}^i_{t}-x_j)^2+(\hat{\mu}^i_{t}-x^\dagger)^2}{N^i_t+1-1} \\ &=\frac{\sum_{j=1}^{N^i_t} x^2_j+(x^\dagger)^2-(N^i_t+1)(\hat{\mu}^i_{t})^2}{N^i_t+1-1} \\
    & \quad = \frac{N^i_t-1}{N^i_t} (\hat{s}^i_{t})^2+\frac{(x^\dagger)^2-(\hat{\mu}^i_{t})^2}{N^i_t}, \quad i \in \{0,1\}~.
\end{align*}

After finding the above estimates of the mean and variance of the truncated distribution, we need to estimate the mean and variance of the \emph{full} underlying distribution. We first note that given our choice of bounds UB and LB, the mean of the underlying distribution is (assumed to be) the same as that of the truncated distribution. To find the untruncated variance for the full distribution, we use the following relation between the variances of truncated and untruncated Gaussian distributions:
\begin{align*}
    Var(x|a\leq x\leq b) =s^2 = \sigma^2 \Bigg[1+\frac{\alpha\phi(\alpha)-\beta\phi(\beta)}{\Phi(\beta)-\Phi(\alpha)}-(\frac{\phi(\alpha)-\phi(\beta)}{\Phi(\beta)-\Phi(\alpha)})^2\Bigg]
\end{align*}
where $\alpha =\frac{a-\mu}{\sigma}$, $\beta = \frac{b-\mu}{\sigma}$, $\phi(x)=\frac{1}{\sqrt{2\pi}} e^{-\frac{1}{2}x^2}$ and $\Phi(x)=\frac{1}{2}(1+erf(\frac{x}{\sqrt{2}}))$. In our algorithm, $a=UB$ and $b=\theta$ for $i=1$, and $a=\theta$ and $b=LB$ for $i=0$. We note that in both cases, we can drop the third term in the above formula since based on our algorithm, $a,b$ are symmetric around the mean value, so that $\phi(\alpha)=\phi(\beta)$. We solve the above equations to find $\hat{\sigma}^i_t$ from the truncated estimates $\hat{s}^i_t$. 

\begin{figure}[t]
	\centering
	\subfigure[Debiasing the means.]
	{
		\includegraphics[width=0.4\textwidth]{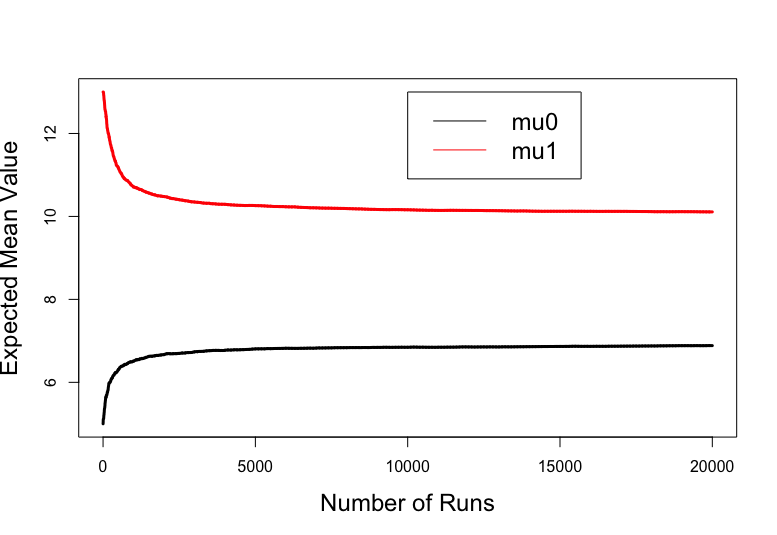}
		\label{fig:fix-mu}
	}
	\subfigure[Debiasing the variances.]
	{
		\includegraphics[width=0.4\textwidth]{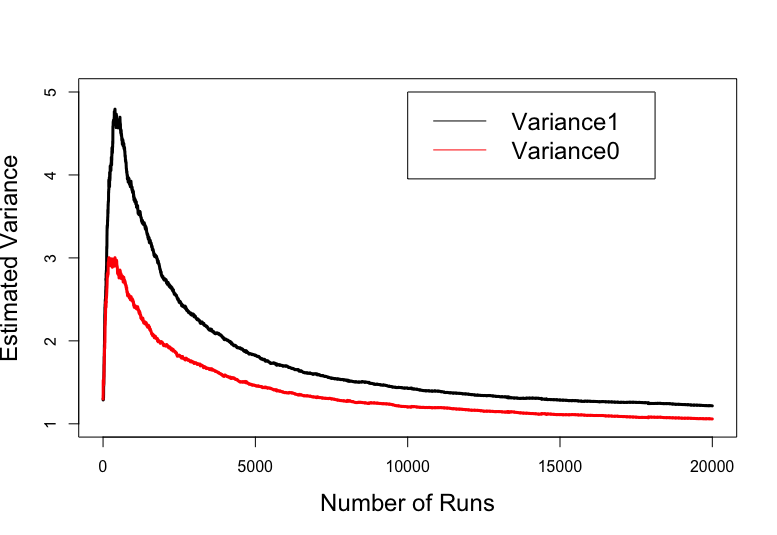}
		\label{fig:fix-sigma}
	}
	\caption{Debiasing algorithm when both mean and variance of a Gaussian distribution are incorrectly estimated. The true underlying distributions are $f^1\sim N(10,1)$ and $f^0\sim N(7,1)$, and the initial estimates are $\hat{f}^1_0\sim N(13,1.3)$ and $\hat{f}^0_0\sim N(5,1.3)$. The algorithm corrects both biases in the long run.}
	\label{fig:two-parameter-debiasing}
\end{figure}

Figure~\ref{fig:two-parameter-debiasing} shows that the debiasing algorithm with the update procedures described above can debias both parameters in the long run. We do observe that the debiasing of the variance initially increases its error. This is because, initially, when observing samples outside of its believed range (due to a combination of incorrectly estimated means and variances), the algorithm increases its estimates of the variance to explain such samples. However, as the estimate of the mean is corrected, the variance can be reduced as well and become consistent with the collected observations. Ultimately, both parameters will be correctly estimated.  

%% file: discussion.tex
\section{More on Limitations, Extensions, and Social Impacts}\label{sec:extensions}

\paragraph{Limitations of Assumption~\ref{as:distributions}, and its extensions.} Our work, similar to the literature on multi-armed bandit learning, focuses on learning of a single unknown parameter (Assumption~\ref{as:distributions}). This assumption has also been adopted in prior work on studying biases in adaptively collected data~\citep{nie2018adaptively}. In addition, our problem is also similar to the commonly studied problem of estimating a \emph{location parameter} of a probability distribution in statistics. Despite the commonality, this assumption entails parametric knowledge of the underlying distribution with the other parameters such as variance or spread being known (or if unknown, immaterial) to the algorithm. This may be a limit in the applicability of our algorithm.  

As a first extension, in Appendix~\ref{app:two_parameters}, we have extended our algorithm and shown that it can debias the data in Gaussian distributions when both the mean and the variance of the distribution are unknown. Our algorithm still uses the same adaptive and bounded exploration mechanism outlined in Algorithm~\ref{def:db-alg}, and the LB set by Definition~\ref{def:UB-LB}. In particular, the bounds are still chosen so as to balance the estimated reference point (i.e. location parameter) of the distribution around its true parameter once correctly estimated; nonetheless, we show the data collected in this way can also be used to debias the variance. 

One potential drawback of this extension is that the debiasing of the variance is not \emph{robust}, in that although the algorithm debiases both parameters in the long-run, it initially {increases} the error in the variance estimate. This is because, initially, when observing samples outside of its believed range (due to a combination of incorrectly estimated means and variances), the algorithm increases its estimates of the variance to explain such samples. However, as the estimate of the mean is corrected, the variance can be reduced as well and become consistent with the collected observations. The extension of our bounded exploration and update procedure to account for both mean and variance debiasing and use of more robust estimators remain as an interesting direction of future work. 

Alternatively, the algorithm could use an input analyzer or distribution fit estimators of a given software to fit a desired distribution to the dataset available at the end of each round, by fixing the known parameters and fitting for the unknown parameters. Exploring this option, especially when additional parameters from the distribution are unknown, remains a direction of future research. 

\paragraph{One-dimensional features and threshold classifiers.} Our current analytical results have focused on one-dimensional feature data and threshold classifiers. Our experiments have also considered the use of our \texttt{active debiasing} algorithm on data with multi-dimensional features by performing a dimension reduction (e.g., as shown in the Adult dataset experiment). 

In general, such dimension reductions may lead to loss of information. To evaluate this in our setting, we run experiments on the Adult dataset with classifiers trained with and without performing dimension reduction. The experiments show only minimal loss in accuracy, with results as follows: 

$\bullet$ For the classifier trained through logistic regression without performing dimension reduction: the overall accuracy is 82.96\%, and the accuracy for advantaged (41762 samples) and disadvantaged (7080 samples) groups are 82.13\% and 88.26\%, respectively.

$\bullet$ For the classifier trained through logistic regression after performing dimension reduction: The overall accuracy is 78.07\%, and the accuracy for the advantaged and disadvantaged groups are 76.95\% and 84.69\%, respectively.

While this remains as a relative small ($\sim$ 5\%) loss in performance, better dimension reduction approaches, or debiasing algorithms that do not require feature dimension reduction, remain as interesting directions of future work. 

\paragraph{Extensions of our analytical results.} We conjecture that Theorem~\ref{thm:debiasing} on the performance of \texttt{active debiasing} can be extended to distributions beyond unimodal distributions. Further, the analytical study of weighted regret of our algorithm, and comparison against the regret incurred by our two baselines, which we have observed numerically in Section~\ref{sec:simulations}, remain as main directions of future work. 
As another future extension, we also propose considering a mixture of normal distributions with varying means and fixed covariances as the assumed form of the underlying distributions that are to be debiased. Such mixtures can be used to approximate other smooth probability density functions. An extension of our analytical results from unimodal distributions to these mixtures, together with extensions of our experimental results, could also address our algorithm's single unknown parameter limitation. 

\paragraph{Potential (negative) social impacts.} We have elaborated on the interplay of our algorithm with fairness constraints in Section~\ref{sec:two-group} as well as in our experiments. Accounting for the different impact of fairness constraints on the speed of debiasing of different groups may be a consideration when choosing to supplement debiasing efforts with fairness constraints, or vice versa. 

Also, a limitation of our algorithm for groups with smaller representation is discussed at the end of Section~\ref{sec:simulations}. In particular, we observed that in the \emph{Adult} dataset, as limited data is available on qualified, disadvantaged agents, the estimates on this population is not fully debiased despite the other estimates having been almost debiased. In other words, our debiasing algorithm is most effective on obtaining correct estimates on populations with sufficiently high representation. This may still be indirectly beneficial to the underrepresented populations, as by having better estimates on the represented population, the algorithm can better assess and impose fairness constraints.

%% file: related-additional.tex
\section{Additional and Detailed Related Work}\label{app:related-additional}

\textbf{Data debiasing with censored and costly feedback.} Our paper is most closely related to the works of \citep{ensign2018runaway,bechavod2019equal,kilbertus2020fair,blum2020recovering,jiang2020identifying}, who have investigated the impacts of data biases on (fair) algorithmic decision making. 
\citet{ensign2018runaway}'s work was one of the earliest to identify the feedback loops between predictive algorithms and biases in the collected data; we investigate similar feedback loops, but are primarily focused on debiasing data, as well as the impact of fairness-constrained learning. 
\citet{bechavod2019equal} and \citet{kilbertus2020fair} study fairness-constrained learning in the presence of censored feedback. While these works also use exploration, the form and purpose of exploration is different: the algorithm in \citep{bechavod2019equal} starts with a pure exploration phase, and subsequently explores with the goal of ensuring the fairness constraint is not violated; the stochastic (or exploring) policies in \citep{kilbertus2020fair} conduct (pure) exploration to address the censored feedback issue. In contrast, we start with a {biased} dataset, and conduct \emph{bounded} exploration with the goal of data debiasing while accounting for the costs of exploration; fairness constraints may or may not be enforced separately and are orthogonal to our debiasing process. {As shown in Section~\ref{sec:simulations}, such pure exploration processes incur higher exploration costs than our proposed bounded exploration algorithm.} 

A number of other works, including  \citep{deshpande2018accurate,nie2018adaptively,neel2018mitigating,wei2021decision} have, similar to our work, explored the question of biases induced by a decision rule on data collection, particularly when feedback is censored. \citet{deshpande2018accurate} study inference in a linear model with adaptively collected data; in contrast to our proposed method, their work focuses on debiasing of an estimator, rather than modifying the decision rule used to collect the data. \citet{nie2018adaptively} study the problem of estimating statistical parameters from adaptively collected data. Their proposed adaptive data collection method, which also similar to ours (Assumption~\ref{as:distributions}) is used for single-parameter estimation, is one of online debiasing; our proposed data collection methods however differ: we propose a \emph{bounded} exploration strategy, which accounts for the risks of exploration decisions and limits the depth of exploration; this method of exploration is different from the random exploration used to collect the data in~\citep{nie2018adaptively}, to which their proposed debiasing algorithms based on data splitting and modified maximum likelihood estimators are applied. 

While \citep{deshpande2018accurate,nie2018adaptively} propose ex-post methods for debiasing adaptively collected data, \citet{neel2018mitigating} consider an adaptive data gathering procedure, and show that no debiasing will be necessary if the data is collected through a differentially private method. We similarly propose a debiasing algorithm that adaptively adjusts its data collection procedure, but unlike \citep{neel2018mitigating}, account for the costs of exploration in our data collection procedure. The recent work of \citet{wei2021decision} studies data collection in the presence of censored feedback, and similar to our work, accounts for the cost of exploration in data collection, by formulating the problem as a partially observable Markov decision processes. Using dynamic programming methods, the data collection policy is shown to be a threshold policy that becomes more stringent (in our terminology, reduces exploration) as learning progresses. Our works are similar in that we both propose using adaptive and cost-sensitive exploration, but we differ in the problem setup and our analysis of the impact of fairness constraints. More importantly, in contrast to both \citep{neel2018mitigating,wei2021decision}, our starting point is a \emph{biased} dataset (which may be biased for reasons other than adaptive sampling in its collection); we then consider how, while attempting to debias this dataset by collecting new data, any {additional} adaptive sampling bias during data collection should be prevented. 

\textbf{Interplay of fairness criteria and data biases.} Our analysis in Section~\ref{sec:two-group}, similar to those of \citet{blum2020recovering} and \citet{jiang2020identifying}, also considers the interplay between algorithmic fairness rules and data biases. \citet{blum2020recovering} show that certain fairness constraints can \emph{themselves} be interpreted as enabling debaising of the underlying estimates. Also, both works 
study data bias arising due to the \emph{labeling} process and propose reweighting techniques to address it. Our work differs from these from two main aspects. First, we model biases as changes in feature-label distributions, in contrast to the assumption of noisy labels in these works. Second, we introduce a statistical debiasing technique based primarily on exploration, which is orthogonal to the social debiasing achieved through fairness constraints. Our proposed model and algorithm therefore complement these works. 

\textbf{Relation to the bandit learning literature.} More broadly, our work is related to the literature on Bandit learning and its study of exploration and exploitation trade-offs, where adaptively adjusted exploration decisions play a key role in allowing the decision maker to attain new information, while at the same time using the collected information to maximize some notion of long-term reward. In particular, bandit exploration deviates from choosing the current best arm in several ways: randomly as in $\epsilon$-greedy, by some form of highest uncertainty as in UCB, by importance sampling approaches as in EXP3, etc. A key difference of our work with these existing approaches is our choice of \emph{bounded} exploration, where the bounds are motivated by settings in which the cost of wrong decisions increase as samples further away from the current decision threshold are admitted. In that sense, our proposed approach can be viewed as a bounded version of $\epsilon$-greedy; we refer to the non-bounded version of $\epsilon$-greedy in our setting as \texttt{pure exploration}, and show that our proposed algorithm can achieve lower weighted regret (one that accounts for the cost of wrong decisions) than \texttt{pure exploration}. 

\textbf{Long-term fairness and bias in algorithmic decision making:} The majority of works on fair algorithmic decision making have focused on achieving fairness in a one-shot setting (i.e. without regards to the long-term effects of the proposed algorithms); see e.g.~\citep{hardt2016equality,dwork2012fairness,corbett2017algorithmic}. Some recent works have studied long-term impacts of fairness on disparities, group representation, and strategic manipulation of features, as a result of adopting fairness measures~\citep{liu2018delayed,zhang2019group,liu2020disparate}. Our work contributes to this line of research, by analyzing the long-term effects of imposing fairness constraints on data collection and debiasing efforts. 

\textbf{Biases in adaptively collected data:} A few other works have, similar to our work, explored the question of biases induced by a decision rule on data collection. \citet{deshpande2018accurate} study inference in a linear model with adaptively collected data; in contrast to our proposed method, their work focuses on debiasing of an estimator, rather than modifying the decision rule used to collect the data.  \citet{nie2018adaptively} study the problem of estimating statistical parameters from adaptively collected data. Their proposed adaptive data collection method, which also similar to ours (Assumption~\ref{as:distributions}) is used for single-parameter estimation, is one of online debiasing; our proposed data collection methods however differ. In particular, our focus is on accounting for multiple subgroups as well as fairness considerations. More importantly, we propose a \emph{bounded} exploration strategy, which accounts for the risks of exploration decisions and limits the depth of exploration; this method of exploration is different from the random exploration used to collect the data in~\citep{nie2018adaptively}, to which their proposed debiasing algorithms based on data splitting and modified maximum likelihood estimators are applied.  

\textbf{Active learning}: Our work is also related to the active learning literature. \citet{balcan2007margin} study the sample complexity of labeled data required for active learning, \citet{kazerouni2020active} propose an algorithm involving exploration and exploitation-based adaptive sampling, verifying it using simulations, \citet{abernethy2020active} propose an active sampling and re-weighting technique by sampling from the worst off group at each step with the goal of building a computationally efficient algorithm with strong convergence guarantees to improve the performance on the disadvantage (highest loss) group while satisfying the notion of min-max fairness, \citet{noriega2019active} propose an adaptive fairness approach, which adaptively acquires additional information according to the needs of different groups or individuals given information budgets, to achieve fair classification. Similar to the approaches of these papers, we also compensate for adaptive sampling bias through exploration (by admitting individuals who would otherwise be rejected). Comparing to all these literature, we start with a biased dataset, and we primarily focus on recovering the true distribution by bounded exploration, accounting for the cost of exploration, avoiding the adaptive sampling bias, and consider fairness issues as orthogonal to our data collection procedure (and as such, can apply our procedure to debiasing the estimates on a single group). 

\textbf{Selective labeling bias}: \citet{lakkaraju2017selective} address the problem of evaluating the performance of predictive model under the selective labeling problem. They propose a contraction technique to compare the performance of the predictive model and human judge while they are forced to have the same acceptance rate. \citet{de2018learning} study the problems arising due to selective labeling. Similar to us, they propose a data augmentation scheme by adding more samples that would be more likely rejected (we refer to this as exploration) to correct the sample selection bias. Their proposed data augmentation technique is similar to our bounded exploration, but it differs in its selection of samples in that it adds samples that would be more likely to be rejected. 

\textbf{Fair learning}: \citet{kallus2018residual} show that residual unfairness remains even after the adjustment for fairness when policies are learned from a biased dataset. They propose a re-weighting technique (similarly, re-weighing ideas are explored in \citep{blum2020recovering} and \citep{jiang2020identifying}) to solve the residual unfairness issue while accounting for the censoring/adaptive sampling bias. 

\textbf{Online mean estimation:} Compared to this literature, the main technical challenges of our proposed bounded exploration in online mean estimation is that it involves evaluating the behavior of statistical estimates based on data collected from a truncated distribution with \emph{time-varying} truncation. More specifically, our data collection interval is bounded and truncated (which has been considered in some prior work on distribution/mean estimation as well, e.g., \citet{lai1991estimating}) but our exploration interval $[LB_t, \infty)$ is itself adaptive (which we believe is the main new aspect) and is what has motivated our analysis in a finite sample regime in Theorem 3. Our focus on the interplay of fairness constraints with online estimation efforts (Proposition 1) is also new compared to this existing literature.

\textbf{Performative prediction:} Finally, the recent line of work on performative prediction proposed by \citet{perdomo2020performative}  also considers the effects of algorithmic decisions on the underlying population's features-label distributions. In particular, the choice of the ML model can cause a shift in the data distribution, and the goal of this work is to identify the stable ML model parameter that is attained at a fixed point of the algorithm-population interactions. In contrast to this goal, our focus is on \emph{pre-existing} and unchanging distribution shifts in the data, which our \texttt{active debiasing} algorithm aims to correct over time. Therefore, our algorithm could be considered as a debiasing method to be used when such performative shifts are present in the data, but are unaccounted for: If distribution shifts happen relatively slower than our debiasing algorithm's convergence speed, our \texttt{active debiasing} could be used to recover correct estimates of the underlying distribution, the estimates of which might have been biased due to performative distribution shifts.

%% file: pseudocode.tex
\SetKwComment{Comment}{/* }{ */}
\RestyleAlgo{ruled}
\begin{algorithm}
\caption{Adaptive Algorithm on Real Data}\label{alg:two}

\textbf{Input}: fairness constraint $\mathcal{C}(\theta_{a,t=0}, \theta_{b,t=0})$, initial fit portion $P$\\
\KwResult{Fair classifier $\theta_{g,t}$}
$t \gets 0, \epsilon_{g,t=0} \gets 1$\\
$\hat{\med}^y_{g,t=0} \gets (\hat{F}^{y}_{g,t=0})^{-1}(\alpha'^y_g)$ for $y \in \{0,1\}$\\
$LB_{g, t=0} \gets (\hat{F}^{0}_{g,t=0})^{-1}(2\hat{F}^{0}_{g,t=0}(\hat{\med}^0_{g,t=0})-\hat{F}^{0}_{g,t=0}(\theta_{g,t=0}))$\\
\While{$i \leq N$}{ 
    $\text{Data\_trun}^y_g$ = [ ]\\
    $\text{portion\_left}^y_g$ = $\hat{F}^y_{g,t}(LB_{t} \leq x \leq \alpha'^y_g)/\hat{F}^y_{g,t}(LB_{t} \leq x)$\\
    $k \gets 0$\\
  \While{$k \leq S$ and $i \leq N$}{
    \For{$g \in G = \{a,b\}$}{
        \uIf{$\theta_{g,t} \leq x$}
            {$\text{Decision} \gets 1$ (accept)}
        \uElseIf{$LB_{g,t} \leq x \leq \theta_{g,t}  \text{ and } rand() \leq \epsilon_{g, t}$}
        {$\text{Decision} \gets 1$ (accept)}
        \uElseIf{$x \leq LB_{g,t}$}
        {$\text{Decision} \gets 0$ (reject)}
        \Else
        {$\text{Decision} \gets 0$ (reject)}
        \textbf{Append} $x$ into $\text{Data}^y_g$ if accepted, and \textbf{append} $x$ into $\text{Data\_trun}^y_g$ only if $x \in (LB_{g, t}, \theta_{g,t})$, or $(\theta_{g,t},\infty)$ with probability $\epsilon_{g,t}$   
    }
  $i \gets i + 1, k = \min(\text{len}(\text{Data\_trun}^y_a),\text{len}( \text{Data\_trun}^y_b))$
  }
  $t \gets t + 1$\\
  $\hat{\med}^y_{g,t} \gets quantile(\text{Data\_trun}^y_g, \text{portion\_left}^y_g)$  \Comment*[r]{Update reference value using all collected samples from the batch}
  \textbf{Map} back from $\hat{\med}^y_{g,t}$ to the single unknown parameter in the estimated distribution $\hat{f}^y_{g,t}$\\
  \textbf{Retrain} the classifier, output new threshold $\theta_{g,t}$ and $LB_{g,t}$ \Comment*[r]{Update classifier using all collected samples so far}
  $\textbf{Update } \epsilon_{g,t}$  \Comment*[r]{Can be FSR or adaptive}
}
\end{algorithm}

%% file: Proof_Thm1.tex
\begin{proof} 
We detail the proof for label 0 estimates $\hat{\parameter}^0_t$, and discuss two cases while assuming (wlog) that the unknown $\parameter^y$ being estimated is the distribution's mean. First, if $\hat{\omega}^0_t$ is overestimated, i.e. $\hat{\omega}^0_t > \omega^0$. Note that we have $\theta_t \geq \hat{\omega}^0_t$. Then, as only agents with $x^\dagger\geq \theta_t$ are admitted, $\hat{\omega}^0_t$ may only be updated to stay the same or increase. Therefore, $\hat{\omega}^0_t$ will remain overestimated. 

Next consider the case that $\hat{\omega}^0_t$ is underestimated, $\hat{\omega}^0_t< \omega^0$. From $t$ on, consider the $T \gg t$ next steps. First, since each observation is independently drawn, we know that at time $t' = t,...,t+T$, $x_{t'} - \mathbb E[X|X \geq \theta_{t'}]$ forms a martingale; this is because of the independence of $x_{t'}$ and $\theta_{t'}$ when conditioned on the historical information, as well as the fact that $\mathbb E[x_{t'}] = \mathbb E[X|X \geq \theta_{t'}]$. 

By definition of $\omega^0$, we also know that $\sum_{t'=t}^T \mathbb E[X|X \geq \theta_{t'}]  > T \cdot \omega^0$. Denote the gap by $\Delta:=\frac{\sum_{t'=t}^T \mathbb E[X|X \geq \theta_{t'}]}{T} -   \omega^0$. Therefore using the Azuma-Hoeffding inequality we have 
\begin{align*}
    \mathbb P\Big(\sum_{t'=t}^T x_{t'} - \sum_{t'=t}^T \mathbb E[X|X \geq \theta_{t'}] \leq \delta \Big) \leq e^{\frac{-2\delta^2}{T-t+1}},
\end{align*}
for any $\delta < 0$. Letting $\delta = -\Delta \cdot (T-t+1)$, the above can be re-written as 
\begin{align*}
     \mathbb P&(\tfrac{1}{T-t+1} \sum_{t'=t}^T x_{t'} > \omega^0) >  1- e^{\left(-2\Delta^2(T-t+1)\right)} \underbrace{\rightarrow 1}_{T \rightarrow \infty}
\end{align*}
This proves that with high probability the mean of the new samples is higher than $\omega^0$. Therefore, at some time $T$ that is significantly higher than $t$, the new estimate $\hat{\omega}^0_T$ will be similar to $\frac{1}{T-t+1}\sum_{t'=t}^T x_{t'}$, which is higher than the true $\omega^0$. From our arguments for the overestimated case, from this point on, $\hat{\omega}^0_t$ will stay overestimated. The proof for $\hat{\omega}^1_t$ is similar. 
\end{proof}

%% file: full_proof_active_debiasing.tex
\begin{proof}
We detail the proof for debiasing $\hat{f}_t^0$ (which happens using $x^\dagger \geq \text{LB}_t$ and $y^\dagger=0$); the proof for $\hat{f}_t^1$ is similar. 

\textbf{Part (a).} In time step $t+1$, with the arrival of a batch of $N_{t+1}$ samples in $[\text{LB}_t, \infty)$, the current estimate $\hat{\parameter}^0_t$ will be updated to $\hat{\parameter}^0_{t+1}$ based on the proportion of $\hat{\parameter}^0_t$ in the existing data. Denote the current left portion in $(\text{LB}_t, \hat{\parameter}^0_t)$ as $p_1 := \frac{\hat{F}^0(\hat{\parameter}^0_t) -\hat{F}^0(\text{LB}_t)}{\hat{F}^0(\theta_t)-\hat{F}^0(\text{LB}_t)}$. Based on Definition ~\ref{def:UB-LB}, we can also obtain the portion in $(\hat{\parameter}^0_t, \theta_t)$ denoted as $p_2 := \frac{\hat{F}^0(\theta_t) - \hat{F}^0(\hat{\parameter}^0_t)}{\hat{F}^0(\theta_t)-\hat{F}^0(\text{LB}_t)} = p_1$. We consider the following cases:

\underline{Case 1 (Perfectly estimated)}: $\hat{\parameter}_t^0=\parameter^{0}$. When the estimates are perfectly estimated, we will have both $\theta_t$ and $\text{LB}_t$ perfectly estimated too. Hence, we have $F^0(\theta_t) - F^0(\hat{\parameter}_t^0) = F^0(\hat{\parameter}_t^0) - F^0(\text{LB}_t)$ such that $p_1 = p_2$. Thus, $\mathbb{E}[\hat{\parameter}^0_{t+1}]=\parameter^0$. 
Hence, once the parameter is correctly estimated,  $\hat{f}_t^0$ is not expected to shift from $f^0$. 

\underline{Case 2 (Underestimated)}: $\hat{\parameter}_t^0<\parameter^{0}$. Under the unimodel distribution and single parameter assumption, since the arriving batch of data comes from the true distribution, $F^0(\text{LB}_t), F^0(\hat{\parameter}_t^0), F^0(\theta_t)$ will be smaller than $\hat{F}^0(\text{LB}_t), \hat{F}^0(\hat{\parameter}_t^0), \hat{F}^0(\theta_t)$, respectively. Moreover, since we have $\text{LB}_t \leq \hat{\parameter}_t^0 \leq \theta_t \leq \hat{\parameter}^1_t$, {then $F^0(\theta_t) - F^0(\hat{\parameter}_t^0) \geq F^0(\hat{\parameter}_t^0) - F^0(\text{LB}_t)$ such that $p_2 \geq p_1$. Hence, more samples are expected to be observed in range of $(\hat{\parameter}^0_t, \theta_t)$ so that the $\hat{\parameter}^0_t$ is expected to shift up.  Hence, we have $\mathbb{E}[\hat{\parameter}^0_{t+1}] \geq \hat{\parameter}^0_{t}$.} 

\underline{Case 3 (Overestimated)}: $\parameter^{0} <  \hat{\parameter}_t^0$.  Through similar analysis as \underline{Case 2 (Underestimated)}, we can obtain $\mathbb{E}[\hat{\parameter}^0_{t+1}] \leq \hat{\parameter}^0_{t}$.

\textbf{Part (b).} We first show that the converging sequence converges to the true estimates. 

By the construction of the bounds in Definition ~\ref{def:UB-LB}, the estimated parameter $\hat{\parameter}^0_{t}$ is the $\alpha$-th percentile of $\hat{f}_t^0$, the median in the interval $[\text{LB}_t, \theta_t]$ and  some percentile in the interval $[\text{LB}_t, \infty)$; we therefore first find their distribution accordingly. Assume there are $N_t=m+n+1$ points in the interval $[\text{LB}_t, \infty)$ with $m$ and $n$ samples below and above $\hat{\parameter}^0_{t}$ respectively. More specifically, for these $n$ samples, there are $m$ samples between $[\hat{\parameter}^0_{t}, \theta_t]$ and $n-m$ samples above $\theta_t$. Based on the probability distribution of order statistics in $[\text{LB}_t, \theta_t]$, denote three possibilities $X$, $Y$, $Z$ denoting the number of samples below, on, and above the $\hat{\parameter}^0_{t}$, respectively, having probabilities $p= \tfrac{F^0(\hat{\parameter}^0_{t})-F^0(\text{LB}_t)}{F^0(\theta_t)-F^0(\text{LB}_t)}$, $q=\tfrac{f^0(\hat{\parameter}^0_{t})}{F^0(\theta_t)-F^0(\text{LB}_t)}$, and $r=\tfrac{F^0(\theta_t)-F^0(\hat{\parameter}^0_{t})}{F^0(\theta_t)-F^0(\text{LB}_t)}$.  Since the distributions are continuous, the probability of multiple samples being exactly on $\hat{\parameter}^0_{t}$ is zero. Therefore, the pdf of $\hat{\parameter}^0_{t}$ can be found based on the density function of the trinomial distribution: 
{\begin{align}
\mathbb P(\hat{\parameter}^0_{t}=\nu) \mathrm{d}\nu= \frac{(2m+1)!}{m!m!}(\tfrac{F^0(\nu)-F^0(\text{LB}_t)}{F^0(\theta_t)-F^0(\text{LB}_t)})^m(\tfrac{F^0(\theta_t)-F^0(\nu)}{F^0(\theta_t)-F^0(\text{LB}_t)})^m \tfrac{f^0(\nu)}{F^0(\theta_t)-F^0(\text{LB}_t)}\mathrm{d}\nu
\label{eq:median-pdf-trinomial}
\end{align}}
From the above, we can see that the density function of the $\hat{\parameter}^0_{t}$ is a beta distribution with $\alpha=m+1, \beta=m+1$, pushed forward by $H(\nu):=\tfrac{F^0(\nu)-F^0(\text{LB}_t)}{F^0(\theta_t)-F^0(\text{LB}_t)}$; this is the CDF of the truncated $F^0$ distribution in $[\text{LB}_t, \theta_t]$. In other words, using $G$ to denote the Beta distribution's CDF, $\hat{\parameter}^0_{t}$ has CDF $G(H(\nu))$, and by the chain rule, pdf $g(H(\nu))h(\nu)$.  

It is known ~\cite{maritz1978note} that for samples located in the range of $[\text{LB}_t, \theta_t]$, the sampling distribution of the median becomes asymptotically normal with mean $(\parameter^0)'$ and variance $\frac{1}{4(2m+3)H((\parameter^0)')}$, where $(\parameter^0)'$ is the median, the truncated $F^0$ distribution in $[\text{LB}_t, \theta_t]$. If the sequence of $\{\hat{\parameter}^0_t\}$ produced by our \texttt{active debiasing} algorithm converges, by Definition ~\ref{def:UB-LB}, the thresholds $\text{LB}_t$ and $\theta_t$ will converge as well; As $t\rightarrow \infty$, $\epsilon_t \rightarrow 0$, $2m+1\rightarrow\infty$ in this interval, the variance becomes zero, and  $\hat{\parameter}^0_{t+1}\rightarrow (\parameter^0)'$. By Definition ~\ref{def:UB-LB}, it must be that the median $(\parameter^0)'$ of $H$ is equal to $\parameter^0$. Therefore, $\hat{\parameter}^0_{t+1}\rightarrow \parameter^0$.

Lastly, we show that the sequence of estimates $\{\hat{\parameter}^0_t\}$ is a converging sequence. Consider the sequence of estimates $\{\hat{\parameter}^0_t\}$, and separate into the two disjoint subsequences $\{\hat{y}^0_t\}$ denoting the parameters that are underestimated with respect to the true $\parameter^0$, and $\{\hat{z}^0_t\}$ denoting those that are overestimated.

We now show that the sequence of underestimation errors, $\{\Delta^y_t\} : = \{\parameter^0- \hat{y}^0_t\}$ and the sequence of overestimation errors, $\{\Delta^z_t\} : = \{\hat{z}^0_t-\parameter^0\}$, are supermartingales. We detail this for $\{\Delta^y_t\}$. Consider two cases: 
\begin{itemize}
    \item First, assume the update $\hat{y}^0_{t+1}$ is the next immediate update after $\hat{y}^0_{t}$ in the original sequence $\{\hat{\parameter}^0_t\}$; that is, an underestimated $\hat{y}^0_{t}$ has been updated to a parameter that continues to be an underestimate. In this case, by Part (a), $\mathbb{E}[\hat{y}^0_{t+1}|\hat{y}^0_{t}]\geq \hat{y}_t^0$, and therefore, $\mathbb{E}[\Delta^y_{t+1}|\Delta^y_{t}]\leq \Delta^y_t$. 
    \item Alternatively assume $\hat{y}^0_{t+1}$ is not obtained immediately from $\hat{y}^0_{t}$; that is, $\hat{y}^0_{t+1}$ has been obtained as a result of an update from an overestimated parameter. We note that now, $\hat{y}^0_{t+1}\geq \hat{y}_t^0$. This is because either no new estimates have been obtained between $\hat{y}_t^0$ and the true parameter $\parameter^0$ since the last time the parameter was underestimated, in which case, it must be that $\hat{y}^0_{t+1}=\hat{y}_t^0$. Otherwise, a new estimate in $[\hat{y}_t^0, \parameter^0]$ has been obtained, in which case, again, $\mathbb{E}[\hat{y}^0_{t+1}|\hat{y}^0_{t}]\geq \hat{y}_t^0$. In either case, $\mathbb{E}[\Delta^y_{t+1}|\Delta^y_{t}]\leq \Delta^y_t$. 
\end{itemize}

Therefore, by the Doobs Convergence theorem, the supermatingales $\{\Delta^y_t\}$ and $\{\Delta^z_t\}$ converge to random variables $\Delta^y$ and $\Delta^z$. By the same argument as the beginning of the proof of this part, these are asymptotically normal with mean zero and with variances decreasing in the number of observed samples in their respective intervals. Therefore, $\Delta^y\rightarrow 0$ and $\Delta^z\rightarrow 0$ as $N\rightarrow \infty$, and therefore $\{\hat{\parameter}^0_t\}$ converges to $\parameter$. 
\end{proof}

%% file: regret_proof.tex
\begin{proof}
{This proof is based on a reduction from fair-classification to a sequence of cost-sensitive classification problems, as proposed and also used to obtain error bounds in \citet{agarwal2018reductions}, and in learning under source and target distribution mismatches as proposed by \citet{ben2010theory}.  We adapt these to our bounded exploration setting.}
In order to find our algorithm's error bound, we proceed through five steps. The first step is to view each individual update of the fair threshold classifier as a saddle point problem, which can be solved efficiently by the exponentiated gradient reduction method introduced in \citet{agarwal2018reductions}. Second, based on the solution output from the reduction method, we find a bound on the classification error achievable based on data from the biased distributions. Thirdly, using results from \citet{ben2010theory}, we bound the error on the target (unbiased) distribution when the algorithm is obtained from the the biased source domain. Then, we will evaluate the impact of exploration errors made by our debiasing algorithm. To simplify notation, we outline the error incurred on one group and at time $t=0$; the effects of errors on other groups and in the subsequent time steps can be similarly obtained. 

In more detail, our algorithm's error is made up of errors from four different sources, which we characterize step by step. Firstly, we have an \emph{approximately} optimal classifer that is returned by the exponentiated gradient algorithm with suboptimiality level $v$ (step 1). Secondly, we use samples to estimate the distributions and thus have empirical biases (step 2). Thirdly, since we start from biased distributions, there are errors due to the domain mismatches (step 3). Lastly, in order to debias, we explore by admitting samples that would otherwise be rejected, introducing additional errors (step 4). 

Before proceeding, we outline our notation for different forms of data bias. First, note that there is a true underlying distribution for the population of agents to which the classifier is to be applied; we denote this by $\bar{D}$. Our focus in this work is on setting where there are different forms of statistical biases in the training data (e.g. distribution shifts or adaptive sampling biases); denote this statistically biased training data by $\tilde{D}$. Finally, even without distribution shifts or adaptive sampling biases, the classifier has access to a limited, empirically biased subset of this data; we denote the initial statistically and empirically biased data distribution by $\hat{D}$. 

Accordingly, let $\hat{h}^*_{\theta_{g,0}}, \tilde{h}^*_{\theta_{g,0}}, \bar{h}^*_{\theta_{g}}$ be the optimal (fair and error minimizing) classifiers that would be obtained from an initial statistically and empirically biased dataset, only statistically biased dataset, and an unbiased dataset,  respectively. 

\noindent \textbf{Step 1: Approximate solution errors.} We can treat the problem of finding the initial fair classifier from the statistically and empirically biased training data as a saddle point problem. First, let $\tilde{err}(h_{\theta_{g,t=0}})=\displaystyle \mathop{\mathbb{E}}_{(x_i,y_i,g_i)\sim \tilde{D}}\Big[\ell(h_{\theta_{g,t=0}}(x_i,g_i),y_i)\Big]$; this is the true error incurred by a classifier $h_{\theta_{g,t=0}}$ when training data comes from $\tilde{D}$, and is the objective function of the minimization problem. Additionally, we assume throughout that a fairness constraint $|\mathcal{C}(\theta_{a,t}, \theta_{b,t})| \leq \gamma$ has been imposed.  

However, since we do not have the true $\tilde{D}$, and only have access to a limited, empirically biased subset of it $\hat{D}$, we will use the empirical estimates $\hat{err}(h_{\theta_{g,0}}), \hat{\mathcal{C}}(\theta_{a,0}, \theta_{b,0})$ and $\hat{\gamma}$ in the constrained optimization problem of finding the fair, loss-minimizing classifier. 

To capture the fairness constraint, we will introduce Lagrangian multipliers $\lambda_j \geq 0$. This allows us to define the Lagrangian of the optimization problem:
\begin{align*}
\mathcal{L}(h_{\theta_{g,0}},\lambda_j)= \hat{err}(h_{\theta_{g,0}}) + \lambda_1 (\hat{\mathcal{C}}(\theta_{a,0}, \theta_{b,0})-\hat{\gamma}) + \lambda_2 (-\hat{\mathcal{C}}(\theta_{a,0}, \theta_{b,0})-\hat{\gamma})
\end{align*}
{Following the rewriting procedures in \citet{agarwal2018reductions}} and using the exponentiated gradient algorithm, we can obtain a $v$-approximated solution $(\hat{h}_{\theta_{g,0}},\hat{\lambda}_j)$; this is an approximately loss-minimizing fair classifier obtained based on an initial, empirically and statistically-biased training data, and the corresponding Lagrange multipliers of the fairness constraint.  

\noindent \textbf{Step 2: Empirical error bound on the initial biased distribution.} To bound the statistical error, we use Rademacher complexity of the classifier family $\mathcal{H}$ denoted as $\mathcal{R}_n(\mathcal{H})$, where $n$ is the number of training samples. Let $n_{g,t}$ be the number of training samples arriving in round $t$ from agents in group $g$. Initially, we have $n_{g,0}= b^0_{g,0}+b^1_{g,0}$. We also assume that $\mathcal{R}_n(\mathcal{H}) \leq Cn^{-\alpha}$ for some $C\geq 0$ and $\alpha \leq 1/2$. Hence, based on the Theorem 4 in \citet{agarwal2018reductions}, we can find that with probability at least $1 - 4\delta$ with $\delta >0$:
\begin{equation}
    \tilde{err}(\hat{h}_{\theta_{g,0}}) \leq \tilde{err}(\tilde{h}^*_{\theta_{g,0}}) +2v + 4\mathcal{R}_{n_{g,0}}(\mathcal{H}) + \frac{4}{\sqrt{n_{g,0}}} + \sqrt{\frac{2\ln(2/\delta)}{n_{g,0}}} 
    \label{eq:empirical_error}
\end{equation}
In words, this provides a bound on the true error that will be incurred on statistically biased data when using the classifier obtained in step 1 (from statistically and empirically biased data). 

\noindent \textbf{Step 3: Bound of error on different distributions (domains).} Next, we note that there is a mismatch between our current biased training data and the true underlying data distribution. We use results from domain adaptation to bound these errors. 

To bound the error on different distributions, $L^1$ divergence would be a nature measure. However, it overestimates the bounds since it involves a supremum over all measurable sets. As discussed by \citet{ben2010theory}, using classifier-induced divergence ($\mathcal{H}\Delta\mathcal{H}$-divergence) allows us
to directly estimate the error of a source-trained classifier on the target domain by representing errors relative to other hypotheses in the hypothesis class. 

\begin{definition}[$\mathcal{H}\Delta\mathcal{H}$-divergence]
For a hypothesis space $\mathcal{H}$, the symmetric difference hypothesis space $\mathcal{H}\Delta\mathcal{H}$ is
the set of hypotheses:
\[g \in \mathcal{H}\Delta\mathcal{H} \Leftrightarrow g(x) = h(x) \oplus h'(x) \hspace{0.1in} \text{for some } h, h' \in \mathcal{H}\]
where $\oplus$ is XOR function. In other words, every hypothesis $g$ is the set of disagreement between two hypotheses in $\mathcal{H}$. The $\mathcal{H}\Delta\mathcal{H}$-distance is also given by
\[d_{\mathcal{H}\Delta\mathcal{H}}(D,D') = 2 \sup_{h,h' \in \mathcal{H}} \Big|Pr_{x\sim D}[h(x) \neq h'(x)] - Pr_{x\sim D'}[h(x) \neq h'(x)] \Big|\]
\end{definition}

Let $\overline{err}(h)$ be the error made by a classifier $h$ on unbiased data from the true underlying distribution. We bound this error below. 
\begin{lemma}[Follows from Theorem 2 of \citet{ben2010theory}]
Let $\mathcal{H}$ be a hypothesis space. If unlabeled samples are from $\tilde{D}_{g,0}$ and $D_g$ respectively, then for any $\delta \in (0,1)$, with
probability at least $1 - \delta$:
\begin{equation}
    \overline{err}(\hat{h}_{\theta_{g,0}}) \leq \tilde{err}(\hat{h}_{\theta_{g,0}}) + \tfrac{1}{2}d_{\mathcal{H}\Delta\mathcal{H}}(\tilde{D}_{g,0}, D_g)+c(\tilde{D}_{g,0}, D_g) 
    \label{eq:source_target}
\end{equation}
where $\tilde{D}_{g,0}$ and $D_g$ are the joint distribution of labels, and $c(\tilde{D}_{g,0}, D_g) = \min_h \overline{err}(h) + \tilde{err}(h)$.
\end{lemma}

Then, combining equation ~\ref{eq:empirical_error} and ~\ref{eq:source_target}, we can obtain the following expression:
\begin{align*}
    &\overline{err}(\hat{h}_{\theta_{g,0}}) \leq  \tilde{err}(\hat{h}_{\theta_{g,0}}) + \tfrac{1}{2}d_{\mathcal{H}\Delta\mathcal{H}}(\tilde{D}_{g,0}, D_g)+c(\tilde{D}_{g,0}, D_g) \\
    &\qquad \leq  \tilde{err}(\tilde{h}^*_{\theta_{g,0}}) +2v + 4\mathcal{R}_{n_{g,0}}(\mathcal{H}) + \tfrac{4}{\sqrt{n_{g,0}}} + \sqrt{\tfrac{2\ln(2/\delta)}{n_{g,0}}} + \tfrac{1}{2}d_{\mathcal{H}\Delta\mathcal{H}}(\tilde{D}_{g,0}, D_g)+c(\tilde{D}_{g,0}, D_g) \\
    &\qquad \leq  \tilde{err}(\bar{h}^*_{\theta_{g,0}}) +2v + 4\mathcal{R}_{n_{g,0}}(\mathcal{H}) + \tfrac{4}{\sqrt{n_{g,0}}} + \sqrt{\tfrac{2\ln(2/\delta)}{n_{g,0}}} + \tfrac{1}{2}d_{\mathcal{H}\Delta\mathcal{H}}(\tilde{D}_{g,0}, D_g)+c(\tilde{D}_{g,0}, D_g) \\
    &\qquad \leq  \overline{err}(\bar{h}^*_{\theta_{g,0}}) +2v + 4\mathcal{R}_{n_{g,0}}(\mathcal{H}) + \tfrac{4}{\sqrt{n_{g,0}}} + \sqrt{\tfrac{2\ln(2/\delta)}{n_{g,0}}} + d_{\mathcal{H}\Delta\mathcal{H}}(\tilde{D}_{g,0}, D_g)+2c(\tilde{D}_{g,0}, D_g) 
\end{align*}
In words, this provides a bound on the true error that will be incurred on the unbiased data from the underlying population when using the classifier obtained in step 1 (from statistically and empirically biased data). 

\noindent \textbf{Step 4: Exploration errors.} Lastly, in order to reduce the mismatches between the biased training data and the true underlying distribution, our algorithm incurs some exploration errors. Let $n'_{0,g,t}$ and $n'_{1,g,t}$ denote the number of samples from unqualified and qualified group that fall below the threshold $\theta_{g,t}$ in round $t$, respectively. Since in Steps 2 and 3 we already considered the classification errors due to empirical estimation and different distributions, we only consider the additional exploration error introduced with the goal of removing biases. Because of exploration, some samples from the {qualified} group that were rejected previously will now be accepted, which will allow the algorithm to make less errors. On the other hand, some samples from the {unqualified} group that would previously be rejected are now accepted, which will lead to an increase in the errors.

Denote $\epsilon_t$ as the exploration probability at round $t$. The exploration error consists of the errors made on the unqualified group, minus correct decisions made on the qualified group. 
In bounded exploration approach, we introduce a $\text{LB}_t$ to limit the depth of exploration. Therefore, the number of samples that fall into the exploration range will be proportional to $n'_{0,g,t}$ and $n'_{1,g,t}$ based on the location of $\text{LB}_t$. Mathematically, denote $N_{g,t}$ as the net exploration error for group $g$ at round $t$; this is given by:
\[N_{g,t} := \Bigg(\frac{\hat{F}^0_{g,t}(\theta_t)-\hat{F}^0_{g,t}(\text{LB}_t)}{\hat{F}^0_{g,t}(\theta_t)}\epsilon_tn'_{0,g,t} - \frac{\hat{F}^1_{g,t}(\theta_t)-\hat{F}^1_{g,t}(\text{LB}_t)}{\hat{F}^1_{g,t}(\theta_t)}\epsilon_tn'_{1,g,t}\Bigg)\]

\noindent \textbf{Step 5: Errors made over $m$ updates.} We now state the error incurred by our algorithm over $m$ rounds of updates. For a group $g$, combining the four identified sources of error over $m$ updates, we have
{\tiny
\begin{align*}
    \sum_{t=1}^{m}\overline{err}(\hat{h}_{\theta_{g,t}}) \leq \sum_{t=1}^{m} \Big[\overline{err}(\bar{h}^*_{\theta_{g,t}}) + 2v + 4\mathcal{R}_{n_{g,t}}(\mathcal{H}) + \tfrac{4}{\sqrt{n_{g,t}}} + \sqrt{\tfrac{2\ln(2/\delta)}{n_{g,t}}} + N_{g,t} + d_{\mathcal{H}\Delta\mathcal{H}}(\tilde{D}_{g,t}, D_g)+2c(\tilde{D}_{g,t}, D_g) \Big]
\end{align*}
}
Therefore, the error bound for our algorithm over $m$ updates and across two groups $g\in\{a,b\}$ is given by 
{
\begin{align*}
&\text{Err.} = 
    \sum_{t=1}^{m}\Big[\overline{err}(\hat{h}_{\theta_{a,t}}) + \overline{err}(\hat{h}_{\theta_{b,t}}) - \overline{err}(h^*_{\theta_{a,t}}) - 
    \overline{err}(h^*_{\theta_{b,t}}) \Big] \\
    &\qquad \hspace{-0.3in} \leq \sum_{g,t} \Big[ \underbrace{2v}_{\text{$v$-approx.}} + \underbrace{4\mathcal{R}_{n_{g,t}}(\mathcal{H}) + \tfrac{4}{\sqrt{n_{g,t}}} + \sqrt{\tfrac{2\ln(2/\delta)}{n_{g,t}}}}_{\text{empirical estimation}} + \underbrace{N_{g,t}}_{\text{explor.}} + \underbrace{d_{\mathcal{H}\Delta\mathcal{H}}(\tilde{D}_{g,t}, D_g)+2c(\tilde{D}_{g,t}, D_g)}_{\text{source-target distribution}} \Big] 
\end{align*}}
\end{proof}
From the expression above, we can see that the more samples we have, $n_{g,t}$ will increase. Hence, the empirical estimation error will decrease. Moreover, as the mismatch between $\tilde{D}_{g,t}$ and $D_g$ is removed by our debiasing algorithm, the mismatch between the source and target domains will also decrease, decreasing the corresponding error term. In the meantime, when $\epsilon_t$ is set adaptively, the exploration probability becomes smaller as we remove the mismatch. Therefore, the exploration error $N_{g,t}$ will also decrease. Together, these mean that the terms in the summation above decrease as $t$ increases. 

%% file: proof_fairness_debiasing.tex
\begin{proof}

We compare the speed of debiasing through $\mathbb{E}[|\hat{\parameter}^y_t-\parameter^y|]$. Given a fixed $t$, we say the algorithm for which this error is larger has a lower speed of debiasing. In words, the slower algorithm needs to wait for \emph{more} arriving samples before it can reach the same parameter estimation error as a faster algorithm. 

We prove the proposition for the case where the introduction of fairness constraints leads to over-selection of group $g$, i.e., $\theta^{F}_{g,t}<\theta^{U}_{g,t}$. The proofs for the under-selected case are similar. We note that the presence of two different groups only affects the choice of the classifier given the fairness constraints, following which the proof becomes independent of the group label; we therefore drop $g$ in the remainder of the proof.

We detail the proof for the debiasing of $\hat{f}^0_{t}$, which depends on the choice of $\text{LB}_t$ in Definition~\ref{def:UB-LB}, i.e., 
\[\hat{F}^{0}_t(\text{LB}_t) = 2\hat{F}^{0}_t(\hat{\parameter}^0_t)-\hat{F}^{0}_t(\theta_t)~.\]
Since $\theta^{F}_{t}<\theta^{U}_{t}$, this means that $\hat{F}^{0}_t(\theta^{F}_{t})<\hat{F}^{0}_t(\theta^{U}_{t})$, and consequently that $\hat{F}^{0}_t(\text{LB}^F_t)>\hat{F}^{0}_t(\text{LB}^U_t)$, and thus, that $\text{LB}^F_t>\text{LB}^U_t$.

Now, consider the interval $[\text{LB}_t, \max^0]$, with $\max^0$ denoting the maximum of ${f}^0$. Only arrivals of $(x^\dagger, y^\dagger)$, with $y^\dagger=0$, who are admitted in this interval, will result in an update to the estimated median. Since $\text{LB}^F_t>\text{LB}^U_t$, this interval is narrower under the fairness constrained classifier, meaning that it takes more time to meet the batch size requirement under compared $\text{LB}^U_t$ compared to $\text{LB}^F_t$. As detailed in the proof of Theorem~\ref{thm:debiasing} each of these updates will move the estimate in the correct direction, and these estimates converge to the true value in the long-run as more samples become available. Hence, debiasing of $\hat{f}_t^0$ is slower after the introduction of fairness constraints. 

Similar arguments hold for updating $\hat{f}^1_{t}$, which takes samples in $[\text{LB}_t, \max^1]$. When $\text{LB}_t$ increases, it also takes more time for label 1 distribution update. Hence, after the introduction of the constraint, the fairness unconstrained classifier observes a wider range of samples points, including all those observed by the constrained classifier. Therefore, the addition of fairness constraints decreases the speed of debiasing on $\hat{f}_t^1$ as well. 
\end{proof}

%% file: main.bbl
\begin{thebibliography}{43}
\providecommand{\natexlab}[1]{#1}
\providecommand{\url}[1]{\texttt{#1}}
\expandafter\ifx\csname urlstyle\endcsname\relax
  \providecommand{\doi}[1]{doi: #1}\else
  \providecommand{\doi}{doi: \begingroup \urlstyle{rm}\Url}\fi

\bibitem[Abernethy et~al.(2020)Abernethy, Awasthi, Kleindessner, Morgenstern,
  Russell, and Zhang]{abernethy2020active}
J.~Abernethy, P.~Awasthi, M.~Kleindessner, J.~Morgenstern, C.~Russell, and
  J.~Zhang.
\newblock Active sampling for min-max fairness.
\newblock \emph{arXiv preprint arXiv:2006.06879}, 2020.

\bibitem[Agarwal et~al.(2018)Agarwal, Beygelzimer, Dud{\'\i}k, Langford, and
  Wallach]{agarwal2018reductions}
A.~Agarwal, A.~Beygelzimer, M.~Dud{\'\i}k, J.~Langford, and H.~Wallach.
\newblock A reductions approach to fair classification.
\newblock In \emph{International Conference on Machine Learning}, pages 60--69.
  PMLR, 2018.

\bibitem[Balcan et~al.(2007)Balcan, Broder, and Zhang]{balcan2007margin}
M.-F. Balcan, A.~Broder, and T.~Zhang.
\newblock Margin based active learning.
\newblock In \emph{International Conference on Computational Learning Theory},
  2007.

\bibitem[Bechavod et~al.(2019)Bechavod, Ligett, Roth, Waggoner, and
  Wu]{bechavod2019equal}
Y.~Bechavod, K.~Ligett, A.~Roth, B.~Waggoner, and S.~Z. Wu.
\newblock Equal opportunity in online classification with partial feedback.
\newblock In \emph{Advances in Neural Information Processing Systems}, pages
  8974--8984, 2019.

\bibitem[Ben-David et~al.(2010)Ben-David, Blitzer, Crammer, Kulesza, Pereira,
  and Vaughan]{ben2010theory}
S.~Ben-David, J.~Blitzer, K.~Crammer, A.~Kulesza, F.~Pereira, and J.~W.
  Vaughan.
\newblock A theory of learning from different domains.
\newblock \emph{Machine learning}, 79\penalty0 (1):\penalty0 151--175, 2010.

\bibitem[Blum and Stangl(2020)]{blum2020recovering}
A.~Blum and K.~Stangl.
\newblock Recovering from biased data: Can fairness constraints improve
  accuracy?
\newblock In \emph{1st Symposium on Foundations of Responsible Computing (FORC
  2020)}. Schloss Dagstuhl-Leibniz-Zentrum f{\"u}r Informatik, 2020.

\bibitem[{CNBC}(2021)]{avg-CS}
{CNBC}.
\newblock {The average U.S. FICO score is up 8 points from last year.}
\newblock
  \url{https://www.cnbc.com/select/heres-how-the-average-american-increased-their-fico-score-last-year/},
  2021.

\bibitem[Corbett-Davies et~al.(2017)Corbett-Davies, Pierson, Feller, Goel, and
  Huq]{corbett2017algorithmic}
S.~Corbett-Davies, E.~Pierson, A.~Feller, S.~Goel, and A.~Huq.
\newblock Algorithmic decision making and the cost of fairness.
\newblock In \emph{Proceedings of the 23rd acm sigkdd international conference
  on knowledge discovery and data mining}, pages 797--806, 2017.

\bibitem[De-Arteaga et~al.(2018)De-Arteaga, Dubrawski, and
  Chouldechova]{de2018learning}
M.~De-Arteaga, A.~Dubrawski, and A.~Chouldechova.
\newblock Learning under selective labels in the presence of expert
  consistency.
\newblock \emph{arXiv preprint arXiv:1807.00905}, 2018.

\bibitem[Deshpande et~al.(2018)Deshpande, Mackey, Syrgkanis, and
  Taddy]{deshpande2018accurate}
Y.~Deshpande, L.~Mackey, V.~Syrgkanis, and M.~Taddy.
\newblock Accurate inference for adaptive linear models.
\newblock In \emph{International Conference on Machine Learning}, pages
  1194--1203, 2018.

\bibitem[Dressel and Farid(2018)]{dressel2018accuracy}
J.~Dressel and H.~Farid.
\newblock The accuracy, fairness, and limits of predicting recidivism.
\newblock \emph{Science advances}, 4\penalty0 (1):\penalty0 eaao5580, 2018.

\bibitem[Dua and Graff(2017)]{Dua:2019}
D.~Dua and C.~Graff.
\newblock {UCI} machine learning repository, 2017.
\newblock URL \url{http://archive.ics.uci.edu/ml}.

\bibitem[Dwork et~al.(2012)Dwork, Hardt, Pitassi, Reingold, and
  Zemel]{dwork2012fairness}
C.~Dwork, M.~Hardt, T.~Pitassi, O.~Reingold, and R.~Zemel.
\newblock Fairness through awareness.
\newblock In \emph{Proceedings of the 3rd innovations in theoretical computer
  science conference}, 2012.

\bibitem[Ensign et~al.(2018)Ensign, Friedler, Neville, Scheidegger, and
  Venkatasubramanian]{ensign2018runaway}
D.~Ensign, S.~A. Friedler, S.~Neville, C.~Scheidegger, and
  S.~Venkatasubramanian.
\newblock Runaway feedback loops in predictive policing.
\newblock In \emph{Conference on Fairness, Accountability and Transparency},
  pages 160--171. PMLR, 2018.

\bibitem[{Experian}(2020)]{avg-CS-2}
{Experian}.
\newblock {What Is the Average Credit Score in the U.S.?}
\newblock
  \url{https://www.experian.com/blogs/ask-experian/what-is-the-average-credit-score-in-the-u-s/},
  2020.

\bibitem[Hardt et~al.(2016)Hardt, Price, and Srebro]{hardt2016equality}
M.~Hardt, E.~Price, and N.~Srebro.
\newblock Equality of opportunity in supervised learning.
\newblock In \emph{Advances in neural information processing systems}, pages
  3315--3323, 2016.

\bibitem[Jiang and Nachum(2020)]{jiang2020identifying}
H.~Jiang and O.~Nachum.
\newblock Identifying and correcting label bias in machine learning.
\newblock In \emph{International Conference on Artificial Intelligence and
  Statistics}, pages 702--712, 2020.

\bibitem[Kallus and Zhou(2018)]{kallus2018residual}
N.~Kallus and A.~Zhou.
\newblock Residual unfairness in fair machine learning from prejudiced data.
\newblock In \emph{International Conference on Machine Learning}, pages
  2439--2448. PMLR, 2018.

\bibitem[Kazerouni et~al.(2020)Kazerouni, Zhao, Xie, Tata, and
  Najork]{kazerouni2020active}
A.~Kazerouni, Q.~Zhao, J.~Xie, S.~Tata, and M.~Najork.
\newblock Active learning for skewed data sets.
\newblock \emph{arXiv preprint arXiv:2005.11442}, 2020.

\bibitem[Kilbertus et~al.(2020)Kilbertus, Rodriguez, Sch{\"o}lkopf, Muandet,
  and Valera]{kilbertus2020fair}
N.~Kilbertus, M.~G. Rodriguez, B.~Sch{\"o}lkopf, K.~Muandet, and I.~Valera.
\newblock Fair decisions despite imperfect predictions.
\newblock In \emph{International Conference on Artificial Intelligence and
  Statistics}, pages 277--287. PMLR, 2020.

\bibitem[Lai and Ying(1991)]{lai1991estimating}
T.~L. Lai and Z.~Ying.
\newblock Estimating a distribution function with truncated and censored data.
\newblock \emph{The Annals of Statistics}, pages 417--442, 1991.

\bibitem[Lakkaraju et~al.(2017)Lakkaraju, Kleinberg, Leskovec, Ludwig, and
  Mullainathan]{lakkaraju2017selective}
H.~Lakkaraju, J.~Kleinberg, J.~Leskovec, J.~Ludwig, and S.~Mullainathan.
\newblock The selective labels problem: Evaluating algorithmic predictions in
  the presence of unobservables.
\newblock In \emph{Proceedings of the 23rd ACM SIGKDD International Conference
  on Knowledge Discovery and Data Mining}, pages 275--284, 2017.

\bibitem[Lambrecht and Tucker(2019)]{lambrecht2019algorithmic}
A.~Lambrecht and C.~Tucker.
\newblock Algorithmic bias? an empirical study of apparent gender-based
  discrimination in the display of stem career ads.
\newblock \emph{Management Science}, 65\penalty0 (7):\penalty0 2966--2981,
  2019.

\bibitem[Lattimore and Szepesv{\'a}ri(2020)]{lattimore2020bandit}
T.~Lattimore and C.~Szepesv{\'a}ri.
\newblock \emph{Bandit algorithms}.
\newblock Cambridge University Press, 2020.

\bibitem[Liao and Naghizadeh(2022)]{liao2022social}
Y.~Liao and P.~Naghizadeh.
\newblock Social bias meets data bias: The impacts of labeling and measurement
  errors on fairness criteria.
\newblock \emph{arXiv preprint arXiv:2206.00137}, 2022.

\bibitem[Liu et~al.(2018)Liu, Dean, Rolf, Simchowitz, and
  Hardt]{liu2018delayed}
L.~T. Liu, S.~Dean, E.~Rolf, M.~Simchowitz, and M.~Hardt.
\newblock Delayed impact of fair machine learning.
\newblock In \emph{International Conference on Machine Learning}, pages
  3150--3158. PMLR, 2018.

\bibitem[Liu et~al.(2020)Liu, Wilson, Haghtalab, Kalai, Borgs, and
  Chayes]{liu2020disparate}
L.~T. Liu, A.~Wilson, N.~Haghtalab, A.~T. Kalai, C.~Borgs, and J.~Chayes.
\newblock The disparate equilibria of algorithmic decision making when
  individuals invest rationally.
\newblock In \emph{Proceedings of the 2020 Conference on Fairness,
  Accountability, and Transparency}, 2020.

\bibitem[Maritz and Jarrett(1978)]{maritz1978note}
J.~Maritz and R.~Jarrett.
\newblock A note on estimating the variance of the sample median.
\newblock \emph{Journal of the American Statistical Association}, 73\penalty0
  (361):\penalty0 194--196, 1978.

\bibitem[Mehrabi et~al.(2019)Mehrabi, Morstatter, Saxena, Lerman, and
  Galstyan]{mehrabi2019survey}
N.~Mehrabi, F.~Morstatter, N.~Saxena, K.~Lerman, and A.~Galstyan.
\newblock A survey on bias and fairness in machine learning.
\newblock \emph{arXiv preprint arXiv:1908.09635}, 2019.

\bibitem[Neel and Roth(2018)]{neel2018mitigating}
S.~Neel and A.~Roth.
\newblock Mitigating bias in adaptive data gathering via differential privacy.
\newblock In \emph{International Conference on Machine Learning}, pages
  3720--3729. PMLR, 2018.

\bibitem[Nie et~al.(2018)Nie, Tian, Taylor, and Zou]{nie2018adaptively}
X.~Nie, X.~Tian, J.~Taylor, and J.~Zou.
\newblock Why adaptively collected data have negative bias and how to correct
  for it.
\newblock In \emph{International Conference on Artificial Intelligence and
  Statistics}, pages 1261--1269, 2018.

\bibitem[Noriega-Campero et~al.(2019)Noriega-Campero, Bakker, Garcia-Bulle, and
  Pentland]{noriega2019active}
A.~Noriega-Campero, M.~A. Bakker, B.~Garcia-Bulle, and A.~Pentland.
\newblock Active fairness in algorithmic decision making.
\newblock In \emph{Proceedings of the 2019 AAAI/ACM Conference on AI, Ethics,
  and Society}, pages 77--83, 2019.

\bibitem[Obermeyer et~al.(2019)Obermeyer, Powers, Vogeli, and
  Mullainathan]{obermeyer2019dissecting}
Z.~Obermeyer, B.~Powers, C.~Vogeli, and S.~Mullainathan.
\newblock Dissecting racial bias in an algorithm used to manage the health of
  populations.
\newblock \emph{Science}, 366\penalty0 (6464):\penalty0 447--453, 2019.

\bibitem[Patil et~al.(2021)Patil, Ghalme, Nair, and
  Narahari]{patil2021achieving}
V.~Patil, G.~Ghalme, V.~Nair, and Y.~Narahari.
\newblock Achieving fairness in the stochastic multi-armed bandit problem.
\newblock \emph{J. Mach. Learn. Res.}, 22:\penalty0 174--1, 2021.

\bibitem[Perdomo et~al.(2020)Perdomo, Zrnic, Mendler-D{\"u}nner, and
  Hardt]{perdomo2020performative}
J.~Perdomo, T.~Zrnic, C.~Mendler-D{\"u}nner, and M.~Hardt.
\newblock Performative prediction.
\newblock In \emph{International Conference on Machine Learning}, pages
  7599--7609. PMLR, 2020.

\bibitem[Raab and Liu(2021)]{raab2021unintended}
R.~Raab and Y.~Liu.
\newblock Unintended selection: Persistent qualification rate disparities and
  interventions.
\newblock \emph{Advances in Neural Information Processing Systems},
  34:\penalty0 26053--26065, 2021.

\bibitem[Reserve(2007)]{fico-data}
U.~F. Reserve.
\newblock Report to the congress on credit scoring and its effects on the
  availability and affordability of credit.
\newblock
  \url{https://www.federalreserve.gov/boarddocs/rptcongress/creditscore/creditscore.pdf},
  2007.

\bibitem[Schumann et~al.(2019)Schumann, Lang, Mattei, and
  Dickerson]{schumann2019group}
C.~Schumann, Z.~Lang, N.~Mattei, and J.~P. Dickerson.
\newblock Group fairness in bandit arm selection.
\newblock \emph{arXiv preprint arXiv:1912.03802}, 2019.

\bibitem[Slivkins(2019)]{slivkins2019introduction}
A.~Slivkins.
\newblock Introduction to multi-armed bandits.
\newblock \emph{arXiv preprint arXiv:1904.07272}, 2019.

\bibitem[Wang et~al.(2021)Wang, Liu, and Levy]{wang2021fair}
J.~Wang, Y.~Liu, and C.~Levy.
\newblock Fair classification with group-dependent label noise.
\newblock In \emph{Proceedings of the 2021 ACM conference on fairness,
  accountability, and transparency}, pages 526--536, 2021.

\bibitem[Wei(2021)]{wei2021decision}
D.~Wei.
\newblock Decision-making under selective labels: Optimal finite-domain
  policies and beyond.
\newblock In \emph{International Conference on Machine Learning}, pages
  11035--11046. PMLR, 2021.

\bibitem[Zhang et~al.(2019)Zhang, Khaliligarekani, Tekin, and
  Liu]{zhang2019group}
X.~Zhang, M.~Khaliligarekani, C.~Tekin, and M.~Liu.
\newblock Group retention when using machine learning in sequential decision
  making: the interplay between user dynamics and fairness.
\newblock In \emph{Advances in Neural Information Processing Systems}, pages
  15269--15278, 2019.

\bibitem[Zhu et~al.(2021)Zhu, Luo, and Liu]{zhu2021rich}
Z.~Zhu, T.~Luo, and Y.~Liu.
\newblock The rich get richer: Disparate impact of semi-supervised learning.
\newblock In \emph{International Conference on Learning Representations}, 2021.

\end{thebibliography}
